\documentclass{ieeeaccess}
\usepackage{cite}
\usepackage{amsmath,amssymb,amsfonts}
\usepackage{algorithmic}
\usepackage{graphicx}
\usepackage{textcomp}
\usepackage{booktabs}
\usepackage{multirow}
\usepackage[table,xcdraw]{xcolor}
\usepackage{hyperref}
\hypersetup{hidelinks=true}

\begin{document}
\history{Date of publication xxxx 00, 0000, date of current version xxxx 00, 0000.}
\doi{10.1109/ACCESS.2023.0322000}

\title{Cross-corpus Readability Compatibility Assessment for English Texts}
\author{\uppercase{Zhenzhen Li}\textsuperscript{*}, 
\uppercase{Han Ding}\textsuperscript{*}, 
and \uppercase{Shaohong Zhang}\textsuperscript{\dag}}
\address{School of Computer Science and Cyber Engineering, Guangzhou University, Guangzhou, 510006, Guangdong Province, China}
\tfootnote{This work was supported by Guangdong Basic and Applied Basic Research Foundation [No. 022A1515011697], and the funding of Guangzhou education scientific research project [No. 1201730714].}
\corresp{Corresponding author: Shaohong Zhang (e-mail: zimzsh@qq.com).}

\begin{abstract}
Text readability assessment has gained significant attention from researchers in various domains. However, the lack of exploration into corpus compatibility poses a challenge as different research groups utilize different corpora. In this study, we propose a novel evaluation framework, Cross-corpus text Readability Compatibility Assessment (CRCA)\footnote{\label{note1}All demo code for the experiments is publicly available at: \href{https://github.com/zimzsh/Cross-corpus-readability-compatibility-assessment-for-English-texts}{https://github.com/zimzsh/CRCA}. Full raw data can be publicly downloaded upon: \href{https://pan.baidu.com/s/12cc8KDw2FhAoT6UgJECZTw?pwd=1111}{https://pan.baidu.com/s/12cc8KDw2FhAoT6UgJECZTw?pwd=1111}. The two mentioned sets of material are also available on request (email: 1973163532@qq.com)}, to address this issue. The framework encompasses three key components: (1) Corpus: CEFR, CLEC, CLOTH, NES, OSP, and RACE. Linguistic features, GloVe word vector representations, and their fusion features were extracted. (2) Classification models: Machine learning methods (XGBoost, SVM) and deep learning methods (BiLSTM, Attention-BiLSTM) were employed. (3) Compatibility metrics: RJSD, RRNSS, and NDCG metrics. Our findings revealed: (1) Validated corpus compatibility, with OSP standing out as significantly different from other datasets. (2) An adaptation effect among corpora, feature representations, and classification methods. (3) Consistent outcomes across the three metrics, validating the robustness of the compatibility assessment framework. The outcomes of this study offer valuable insights into corpus selection, feature representation, and classification methods, and it can also serve as a beginning effort for cross-corpus transfer learning.
\end{abstract}


\titlepgskip=-21pt

\maketitle

\section{Introduction}
\label{sec:introduction}
\PARstart{W}{ith} the advent of globalization and digitization, English has become a universal language \cite{drubin2012english}. For non-native English learners, reading proficiency is of paramount importance \cite{huckin1995second}. Teachers often use reading exercises to enhance the reading skills of their students, but it is essential to ensure that the texts are suitable for the students' level \cite{o2002teaching}. However, the discrepancy in readability levels among different corpora poses challenges for non-native English learners. Thus, it becomes necessary to investigate the compatibility of English text difficulty across different corpora.

The compatibility of cross-corpus English text difficulty refers to the correlation between the readability assessment results of the same English text in different corpora. Over the past few decades, researchers in natural language processing (NLP) have explored various methods, including statistical language models, feature-based machine learning methods, and state-of-the-art deep neural networks, to assess readability. While Automatic Readability Assessment (ARA) often employs traditional readability formulas, these formulas tend to overlook intricate aspects within the text. Although ARA is typically approached as a supervised learning problem \cite{vajjala2014readability, ma2012ranking}, a consensus on the compatibility of cross-corpus difficulty has yet to be reached. While Lee and Vajjala \cite{lee2022neural} proposed a neural pairwise ranking model that demonstrated promising results in zero cross-language transfer, a comprehensive analysis of cross-corpus compatibility is still lacking. Most studies describing ARA models typically focus on metrics such as classification accuracy, F-score, Pearson correlation, Spearman correlation, and root mean squared error. Although some evaluations have considered multiple corpora, either by training on one corpus and testing on multiple ones \cite{lee2022neural} or training and testing on multiple corpora \cite{vajjala2014readability, xia2019text}, there lacks standardized metrics for tasks involving cross-corpus compatibility. Moreover, although a few studies have explored the application of cross-corpus analysis \cite{azpiazu2019multiattentive, madrazo2020cross}, there is a need for a comprehensive framework to assess compatibility and to compare new corpora with widely used benchmark corpora.

Addressing the aforementioned issues, our work makes a threefold contribution. Firstly, it serves as an exploratory study on cross-corpus readability compatibility. Secondly, three metrics are proposed to evaluate cross-corpus compatibility, which can serve as a basis for validating corpus selection. Thirdly, a systematic framework is established to evaluate and compare the compatibility of new corpora with benchmark corpora. Additionally, our research has implications for applications such as corpus data augmentation and transfer learning. Specifically, in this study, machine learning and deep learning techniques and employed. Three metrics, including Reverse-Jensen-Shannon Divergence (RJSD), Reverse-Rank Normalized Sum of Squares (RRNSS), and Normalized Discounted Cumulative Gain (NDCG), are adopted to assess compatibility. Experimental results demonstrate the effectiveness of our proposed text readability compatibility assessment system for cross-corpus analysis, addressing the limitations of previous studies in this area.

The remainder of this paper is organized as follows. Section 2 provides an overview of related research. Section 3 defines the problem of cross-corpus text readability assessment compatibility. Section 4 introduces the corpora and features used in this study. Section 5 presents a detailed description of the readability compatibility assessment system and model construction. Section 6 analyzes the performance of the model and evaluates cross-corpus compatibility. Finally, we discuss the experimental conclusions of this paper and present future prospects.

\section{Related Work}
ARA has been a cross-disciplinary topic in education, psychology, and computer science for almost a century. At the core of ARA tasks lies the concept of text difficulty, which refers to the level at which learners can read and comprehend text materials \cite{vajjala2021trends}. 

Early methods of evaluating text readability primarily relied on manually designed readability formulas. These formulas offered objectivity, simplicity, and cost-effectiveness in assessing text difficulty. However, Thomas Oakland \cite{oakland2004language} pointed out limitations in readability formulas. For instance, the Flesch-Kincaid formula only considers text features such as sentence length and word difficulty, neglecting other factors that may impact readability. Similarly, Vajjala \cite{vajjala2021trends} highlighted that traditional readability formulas can only be trained and tested on specific domains or text types.

The main research results on the difficulty measurement of English texts have developed hundreds of text readability measurement formulas. Several typical readability measurement formulas are shown in Table \ref{Readability}.

ARA has witnessed the dominance of neural network-based architectures in recent years, following the trend observed in other NLP studies. Researchers, such as Mohammadi and Khasteh \cite{mohammadi2019text} and Meng \cite{meng2020readnet}, have proposed various neural network models for multilingual readability assessment. These models incorporate deep reinforcement learning and hierarchical self-attention based transformer architectures, respectively. Word embedding techniques have been combined with additional attributes, such as domain knowledge and language modeling, to enhance performance \cite{cha2017language, kim2012characterizing}. A wide range of neural architectures, including multi-attention RNN and deep reinforcement learning, have been explored in the pursuit of improved ARA models \cite{deutsch2020linguistic, lee2021pushing}.

While there have been studies examining cross-corpus compatibility, the focus remains limited. For instance, Francois and Fairon \cite{franccois2012ai} utilized sentence alignment methods to construct a parallel corpus of French and English for corpus construction. Xia \cite{xia2019text} leveraged out-of-domain training data to improve performance on limited in-domain data. Azpiazu and Pera \cite{azpiazu2019multiattentive, madrazo2020cross} explored deep learning architectures to investigate multilingual and cross-lingual approaches to ARA. Weiss \cite{weiss2021using} explored the effectiveness of linguistic features across different languages and conducted a zero-shot cross-lingual evaluation between English and German, utilizing an extensive set of linguistic features. However, to the best of our knowledge, there is currently no existing work specifically evaluating the compatibility of cross-corpus readability.

In summary, for readability assessment research, cross-corpus difficulty compatibility is an important problem to be solved. This study utilized six different corpora and employed machine learning and deep learning methods, as well as various feature combinations and evaluation metrics. These efforts greatly advanced research progress in compatibility assessment tasks and also provided broader prospects for the application and development of readability assessment.

\renewcommand{\arraystretch}{1.5} 
\begin{table*}[htp]
  \caption{Traditional readability formula, including Automated Readability Index (ARI), Flesch-Kincaid Grade Level (FKGL), Gunning Fog Index (GFI), SMOG Grading (SMOG), Coleman-Liau Index (CLI), Lesbarhets Index (LIX), and Rate Index (RIX)}
  \label{Readability}
  \centering
  \footnotesize
  \begin{tabular}{ll}
  \toprule[1.5pt]
  Name & Readability Formula  \\
  \midrule
  ARI \cite{senter1967automated}  & $ARI=4.71\left(\frac{{ characters }}{{ words }}\right)+0.5\left(\frac{{ words }}{{ sentences }}\right)-21.43$  \\
  
  FKGL \cite{kincaid1975derivation} & $FKGL=0.39\left(\frac{{words}}{{ sentences }}\right)+11.8\left(\frac{{ Syllables }}{{words }}\right)-15.59$  \\ 
  
  GFI \cite{gunning1952technique} & $GFI=0.4(\left(\frac{words}{sentence}\right)+100\left(\frac{complexwords}{words}\right))$  \\ 
  
  SMOG \cite{mc1969smog}  & $SMOG=\sqrt{complexWords(\frac{30}{sentences})}+3.1291$ \\
  
  CLI \cite{coleman1975computer}  & $CLI=5.89\left(\frac{characters}{words}\right)-30\left(\frac{sentences}{words}\right)-15.8 $\\ 
  
  LIX \cite{bjornsson1983readability}  & $LIX=\frac{words}{sentences}+100(\frac{longwords}{words}) $\\ 
  
  RIX \cite{anderson1983lix} & $RIX=\frac{longwords}{sentences}$ \\
  \bottomrule[1.5pt]
  \end{tabular}
\end{table*}

where, 'characters' is the total number of characters, 'words' is the total number of words, 'sentences' is the total number of sentences, 'complexWords' is the number of complex words, 'longwords' is the number of words longer than 6 characters, 'Syllables' is the total number of syllables, and 'Polysyllables' is the total number of polysyllabic words.

\section{Problem Description}
Cross-corpus difficulty compatibility assessment requires some preliminary knowledge. In this section, we propose two hypotheses based on background knowledge. To make it clearer, the relevant variable attributes in the consistent evaluation task are given as follows :
\begin{itemize}
    \item $L2$ stands for the corpus collection of second language learners;
    \item $S$ stands for the source corpus for model training and difficulty evaluation; 
    \item $T$ stands for the target corpus for model evaluation and testing; 
    \item $f(S,T)$ stands for the cross-corpus difficulty evaluation function, and uses the model trained by the source corpus $S$ to predict the difficulty level of the text of the target corpus $T$; 
    \item $\omega(S,T)$ stands for the compatibility of difficulty levels between the source corpus and target corpus. In particular, due to the particularity of the compatibility assessment process proposed in this paper, $\omega(S,T)$ and $\omega(T,S)$ are not numerically equal.
    
\end{itemize}

\textbf{Assumption 1}: The difficulty level of each text in the corpus can be represented by an integer, and the distance between difficulty levels is ignored. The distance between each difficulty level is equal, as follows:

\begin{equation}
    d(x_i,x_j)=|x_i-x_j|
\end{equation}

where, $d(x_i, x_j)$ stands for the distance between the difficulty levels of text $i$ and text $j$, $x_i$ and $x_j$ stand for the difficulty levels of text $i$ and text $j$, and $|\cdot|$ stands for the absolute value function.

\textbf{Assumption 2}: Text labels between different corpora can be transferred, such as readability formulas or general text difficulty features. Specifically, it is assumed that the difficulty level of each text can be represented by a function $f(\cdot)$:

\begin{equation}
    y_i=f(x_i)
\end{equation}

where, $x_i$ stands for the difficulty level of text $i$ in the target corpus $T$, and $y_i$ stands for the difficulty level of text $i$ in the source corpus $S$. The function $f(\cdot)$ stands for the transformation function from the difficulty level of corpus $T$ to the difficulty level of corpus $S$.

\textbf{Definition (Cross-corpus readability assessment compatibility)}: Assuming that we select two corpora from $L2$, which are respectively used as the source corpus $S$ and the target corpus $T$, with readability difficulty levels of $L_s$ and $L_t$. The assessment model trained by the source corpus $S$ is used to evaluate the target corpus $T$, and the prediction difficulty label $L_{t'}$ is obtained.

\begin{equation}
    \Omega(T,S)=cor(L_t,L_{t'})
\end{equation}

where, $cor(\cdot)$ stands for the assessment method, which is $RJSD$, $RRNSS$ and $NDCG$ in this paper.

The study learns a model $Model(S)$ from the source corpus that can map text to the corresponding difficulty level, and predicts the difficulty of the text in the corpus $T$. Based on the above premises and definitions, this study aims to answer the following two questions: 1) Can similar compatibility assessment results be obtained by using different model training methods and feature combinations? 2) Can the method proposed in this study be universal on different corpora?

Compared with most previous works, this paper answers these questions based on experiments across multiple cross-corpus. Traditional readability formulas cannot reflect the compatibility of readability assessment when evaluating cross-corpus texts. Therefore, this study uses machine learning and deep learning methods to model cross-corpus text.

\section{Data and Features}
\subsection{Datasets}
Our study employs six common datasets for readability research. In the experiment,  all pre-processed samples from the datasets are included. The data set details are shown in Table \ref{tab:Dataset}.

\renewcommand{\arraystretch}{1.2}
\begin{table*}[htp]
\caption{Dataset Information}
\label{tab:Dataset}
\footnotesize
\begin{tabular}{lp{12cm}cl}
\toprule[1.5pt]
Dataset & Description \centering & Volume & References \\
\midrule
CEFR     
& The Common European Framework of Reference for Languages (CEFR) is an internationally recognized standard for describing language proficiency and levels. The standard divides six levels based on language texts of different difficulty levels, from beginner to professional level \cite{ council2001common}. 
& 683  
&\cite{khushik2022syntactic}\cite{huang2023development}\cite{negishi2013progress}   \\
\hline

CLEC 
& Chinese Learner English Corpus (CLEC) is a corpus based on learners' English writing. There are five levels of corpus, including middle school students, college English level 4 and 6, junior and senior professional English \cite{Yang2005corpus}. 
& 4709 
&\cite{chen2017corpus} \cite{he2016computer} \cite{wang2019corpus}\\
\hline

CLOTH 
& CLOTH is a large-scale English cloze test set, which is divided into two grades: junior high school and senior high school \cite{xie2017large}.
& 7129 
&\cite{liu2019neural} \cite{xu2022fantastic} \cite{lv2020pre} \\
\hline

NES 
& Newsela (NES) corpus consists of news articles. It covers articles from grade 2 to grade 12 at different levels, allowing students to practice reading on texts appropriate to their ability level \cite{xu2015problems}. 
& 33006 
&\cite{azpiazu2019multiattentive} \cite{deutsch2020linguistic} \cite{martinc2021supervised} \\
\hline

OSP 
& The OneStopEnglish (OSP) corpus contains three levels of reading difficulty (primary, intermediate, and advanced), with 189 texts at each level \cite{vajjala2018onestopenglish}. & 567 
&\cite{lee2022neural} \cite{martinc2021supervised} \cite{vajjala2021trends} \cite{imperial2021bert} \\
\hline

RACE 
& The Reading Comprehension Dataset From Examination (RACE) is a collection of reading comprehension materials from English exams for middle and high school students in China, divided into two levels based on grades: junior high and senior high \cite{lai2017race}. 
& 27931 
&\cite{chen2018learningq} \cite{xu2022fantastic} \cite{si2020benchmarking} \cite{malmaud2020bridging} \\

\bottomrule[1.5pt]
\end{tabular}
\end{table*}

In general, different datasets have different characteristics and readability systems.
\begin{itemize}
    \item Data Volume: Among the six datasets, NES has the most data with 33,006 records, while RACE has 27,931 records. In comparison, CLOTH and CLEC have relatively few records, which is less than 10,000; and CEFR and OSP have the least amount of data, with 683 and 567 records, respectively.
    \item Text Types: NES and OSP datasets include news articles. CLEC, CLOTH, and RACE contain exam questions, with CLEC being the Chinese English proficiency test data, CLOTH being the cloze test data, and RACE being the reading comprehension data. The CEFR dataset comes from various text types, including online resources, textbooks, social media, and news media, covering language usage scenarios for different topics.
    \item Readability Levels: From the description of the datasets in Table 2, it can be seen that NES has the widest range of readability levels, from 2nd grade to 12th grade, with 11 levels included. CEFR and CLEC are next, each containing 6 and 5 levels respectively; OSP contains 3 levels; while CLOTH and RACE are divided into 2 levels based on junior high and senior high.
\end{itemize}

The datasets used in this study have different difficulty systems. Appendix \ref{appendix:index score} shows the assessment results of traditional readability formulas on the datasets. It can be seen that the data set text has a difficulty level. At the same time, we show the distribution of traditional readability formula scores for difficulty levels in each dataset in Appendix \ref{appendix:Distribution}. 

\subsection{Features}
Feature extraction is a important part of machine learning, which is the process of converting raw data into useful information representations. We extracted features from six datasets respectively, which were used as inputs for subsequent machine learning and deep learning methods. In this study, the readability assessment features used are divided into three categories: lexical features, syntactic features, and semantic features. Appendix \ref{appendix:Feature} shows the feature information.

\section{Model Design}
\subsection{Cross-corpus Assessment System}
To address the problems described in Section 3, this paper proposes a method for cross-corpus assessment, which predict the readability of texts based on the knowledge of  different corpora, so as to evaluate the cross-corpus compatibility.

\begin{figure}[htb]
    \centering
    \includegraphics[width=8.5cm]{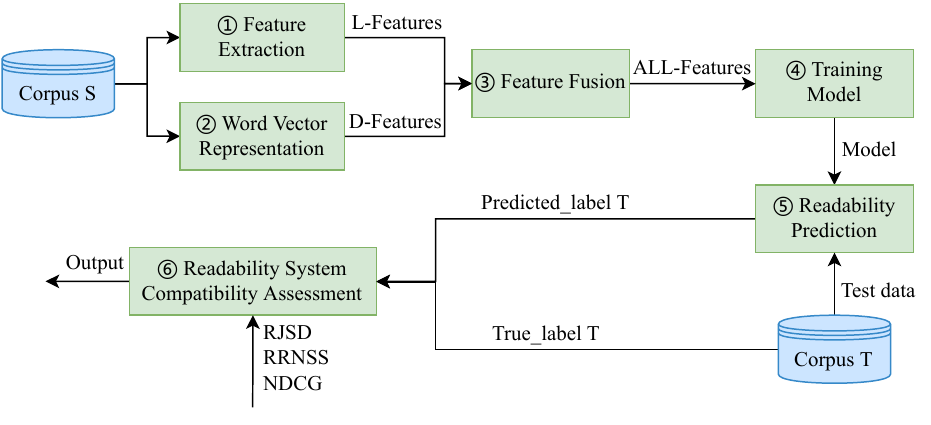}
    \caption{Cross-corpus assessment process}
    \label{fig:Cross-corpus Evaluation}
\end{figure}

The overall framework of our work is shown in Figure \ref{fig:Cross-corpus Evaluation}, which includes six steps for evaluating the compatibility of cross-corpus difficulty system.

\textbf{Step 1. Feature Extraction:} In natural language processing, feature extraction is a crucial process. In this process, we retrieve the most suitable features from the corpus and convert the text into a vector space to optimize the quality and efficiency of the model. In this study, 21 linguistic features were calculated for six corpora, including three levels: lexical, grammatical, and syntactic aspects. The details of the feature information are shown in Appendix Appendix \ref{appendix:Feature}, collectively referred to as L-Features.

\textbf{Step 2. Word Vector Representation:} When using GloVe for text representation, we extract all possible word combinations from the corpus according to a pre-set window size and count their frequency of occurrence in the text. In this way, a co-occurrence matrix based on word frequency can be obtained. The co-occurrence matrix is then decomposed into two low-dimensional matrices to obtain a vector representation of each word. Therefore, D-Features using GloVe can better represent the semantic relationship between words in terms of text representation, providing a more optimized way of representing text.

\textbf{Step 3. Feature Fusion:} In order to comprehensively utilize the multi-level information of text, we fuse traditional text features with word vector representations. And the new feature representation is used for subsequent model training. Specifically, we convert each sentence in the text data into a corresponding feature vector in the machine learning method, and concatenate the linguistic feature data with the sentence-level feature vectors element-wise to obtain a multi-level text representation. In the deep learning approach, we concatenate the LSTM output through a concatenate layer, process the linguistic features through a dense layer. Finally, the concatenate layer is used to combine both types of features into a new feature representation called ALL-Features for subsequent model training.

\textbf{Step 4. Training Model:} In this study, we trained the models using different learning methods, including XGBoost, SVM, BiLSTM, and Attention-BiLSTM. To explore the impact of different feature combinations on compatibility assessment, we also conducted three sets of experiments based on different feature combinations. The first set of experiments trained traditional machine learning models on the extracted linguistic features to predict text complexity. The second set of experiments evaluated readability using GloVe word vector representations with both machine learning and deep learning methods. The third set of experiments evaluated readability using features obtained by fusing linguistic features and word vectors with both machine learning and deep learning methods. It should be noted that deep learning methods can extract more information from raw text data, helping models learn more complex feature representations and improve generalization ability. In contrast, low-dimensional language features are limited by lack of information richness and representational power, which limit effectiveness in deep learning models. During model training, cross-validation was used to evaluate the performance of the models, and the accuracy value was used as the main metric for model selection and optimization.

\textbf{Step 5. Readability Prediction:} We use the classification model trained in Step 4 to predict the text readability of the target corpus $T$. By obtaining the difficulty labels predicted by models trained on different source corpora for the target corpus, we can further analyze the compatibility assessment of cross-corpus text difficulty. This study trains models using different combinations of learning methods, feature combinations, and source corpora. We found that different learning methods and feature combinations have different effects on cross-corpus compatibility assessment of text difficulty.

\textbf{Step 6. Readability System Compatibility Assessment:} We will evaluate the compatibility between the predicted labels and the real labels of the target corpus in the fifth step. This paper uses the evaluation metrics RJSD, RRNSS and NDCG to measure the compatibility between the corpus difficulty systems, and compares the results of the three methods. 

Through the above steps, we can get the readability assessment results of cross-corpus texts. By comparing the compatibility between the difficulty systems of different corpora, we can further analyze the differences and rules of text difficulty across corpora.

\subsection{Learning Algorithms}
In this study, we employed both machine learning and deep learning approaches with different feature combinations for experimentation. Specifically, the machine learning method utilized language features, GloVe word embeddings, and their fusion features, while the deep learning method only used GloVe word embeddings and fusion features. Additionally, we compared and analyzed the performance of different feature combinations.

\begin{itemize}
    \item XGBoost: We utilized the XGBoost classifier proposed by Chen and Guestrin \cite{cortes1995support} to train a model for text readability assessment.
    \item SVM: Support Vector Machines constructs a hyperplane that separates the data into classes. SVM is efficient for high-dimensional feature spaces \cite{dumais1998inductive, joachims2005text}.
\end{itemize}

\begin{itemize}
    \item Bi-LSTM: The designed Bidirectional LSTM neural network is based on the research conducted out by Sachan \cite{stevens1946theory}. The implemented model architecture is a sequential architecture consisting of an embedding layer, a single bidirectional LSTM layer, a pooling layer, and a fully-connected layer with softmax function for classification.  All hyperparameters are tuned using the development set.
    \item Att-BiLSTM: Attention-based bidirectional long short-term memory is a model that adds an Attention layer on top of the BiLSTM model. In English text, different words contribute differently to the overall semantics of a sentence. Based on the attention mechanism, we assign weights according to the contribution of words to text semantics, so that the classifier pays more attention to semantic information and improves the classification performance of the model.
\end{itemize}

\section{Results and Evaluation}
In this section, we will present the experimental results. Section 6.1 introduces the evaluation metrics used in the study, and Section 6.2 presents the experimental results and analysis.

\subsection{Evaluation Metrics}
\subsubsection{Model Performance Metrics}
In this section, the performance of each classifier are evaluated and related results are shown in Appendix \ref{appendix:Model Performance}. The following four metrics are  used: (a) Accuracy, (b) Precision, (c) Recall, (d) F1. The definitions of the formulas are as follows:

\begin{equation}
    Accuracy=\frac{TP+TN}{TP+TN+FP+FN}
\end{equation}

\begin{equation}
    Precision=\frac{TP}{TP+FP}
\end{equation}

\begin{equation}
    Recall=\frac{TP}{TP+FN}
\end{equation}

\begin{equation}
    F1=\frac{2\times\mathrm{Precision}\times\mathrm{Recall}}{\mathrm{Precision}+\mathrm{Recall}}
\end{equation}

where, true positive (TP) is the number of data classified as positive in the data marked as positive, and true negative (TN) is the number of data classified as negative in the data marked as negative. False positive (FP) is the number of data classified as positive but marked as negative in the data set, and false negative (FN) is the number of data classified as negative but marked as positive in the data set. Accuracy represents the proportion of correct prediction of the model. Precision refers to the proportion of the number of correctly classified texts to the total number of texts classified into different readability levels. Recall refers to the ratio of the number of correct classifications to the total number of texts in each test set. F1 is the average of Precision and Recall.

\subsubsection{Compatibility Assessment Metrics}
To evaluate the compatibility of cross-corpus, we adopted three assessment metrics: Reverse-Jensen-Shannon Divergence (RJSD), Reverse-Rank Normalized Sum of Squares (RRNSS), and Normalized Discounted Cumulative Gain (NDCG). The following describes the detailed definition of indicators.

RJSD is a new metric based on Jensen-Shannon Divergence (JSD) \cite{lin1991divergence}. JSD is a commonly used method for measuring the similarity between probability distributions, proposed by Jensen and Shannon. Its value ranges from 0 to 1, with smaller values indicating greater similarity between distributions. In this study, JSD is adjusted so that the value range of RJSD is [0,1], with larger values indicating greater similarity between distributions. The formula for calculating RJSD is as follows:

\begin{equation}
    RJSD(P,Q)=1-\frac{KL(P|M)+KL(Q|M)}{2}
\end{equation}

where, P and Q stand for the two probability distributions to be compared, and M stands for their average distribution.

RRNSS is a new metric based on Rank Normalized Sum of Squares (RNSS). RNSS is essentially the Root Mean Squared Error (RMSE), and the advantages of RMSE over MAE have been discussed in Chai and Draxler \cite{chai2014root}. In the calculation of RNSS, the elements in the two ranking lists are standardized and converted into a numerical list in the range of [0,1], and then the sum of squared differences and weighted sum are calculated for these two standardized lists to obtain the RNSS value. The calculation formula of the adjusted RRNSS metric in this study is as follows:

\begin{equation}
    \mathrm{RRNSS}=1-\left[1+\left(\frac{\sum_{i=1}^n(A_i-B_i)^2}{n(n+1)(2n+1)/6}\right)^{0.5}\right]^{-1}
\end{equation}

where, $Ai$ stands for the ranking of the $i$ th element in list $A$, and Bi stands for the ranking of the $i$ th element in list $B$. RRNSS has a value range between 0 and 1. The larger the RRNSS value, the closer the sorting result is to the true sorting, and the better the sorting quality.

NDCG is a metric used to evaluate the quality of rankings, taking into account both the relevance of the ranking result and the order in which they are ranked \cite{lee2022neural,busa2012apple}. The calculation method of NDCG is based on the Discounted Cumulative Gain (DCG) metric \cite{jarvelin2008discounted}. DCG is a metric that measures the quality of ranking results and reflects the gap between the ranking results and the ideal ranking results. Its calculation formula is as follows:

\begin{equation}
    DCG_k=\sum\limits_{i=1}^k\frac{rel(i)}{log_2(i+1)}
\end{equation}

where, $rel_i$ represents the relevance score of the $i$ th text, and $k$ is the number of results. In order to eliminate the influence of the number of ranking results, DCG needs to be normalized to obtain the NDCG metric. The formula for calculating NDCG is as follows:

\begin{equation}
    NDCG=\frac{DCG}{IDCG}
\end{equation}

where, IDCG is the DCG value of the ideal ranking result. The value range of NDCG metric is between 0 and 1, and the larger the value, the higher the relevance between the ranking result and the true label.

\subsection{Experimental Results and Analysis}
We conducted readability assessments on six corpora using three experimental combinations, including machine learning+features, machine learning+GloVe word vectors, deep learning+GloVe word vectors, machine learning+fusion features, and deep learning+fusion features. This study used three assessment metrics to evaluate the compatibility of prediction results. Based on the length of this section, we mainly discuss the RJSD assessment metric. The results of RRNSS are shown in Figure \ref{fig:RRNSS (ML+Feature)}, Figure \ref{fig:RRNSS (ML/DL+GloVe)}, and Figure \ref{fig:RRNSS (ML/DL+Fusion Feature)}. The results of NDCG are shown in Figure \ref{fig:NDCG (ML+Feature)}, Figure \ref{fig:NDCG (ML/DL+GloVe)}, and Figure \ref{fig:NDCG (ML/DL+Fusion Feature)}.

\subsubsection{Compatibility Analysis of Using Linguistic Features in Machine Learning}

\begin{figure}[htb]
    \centering
    \includegraphics[width=8.5cm]{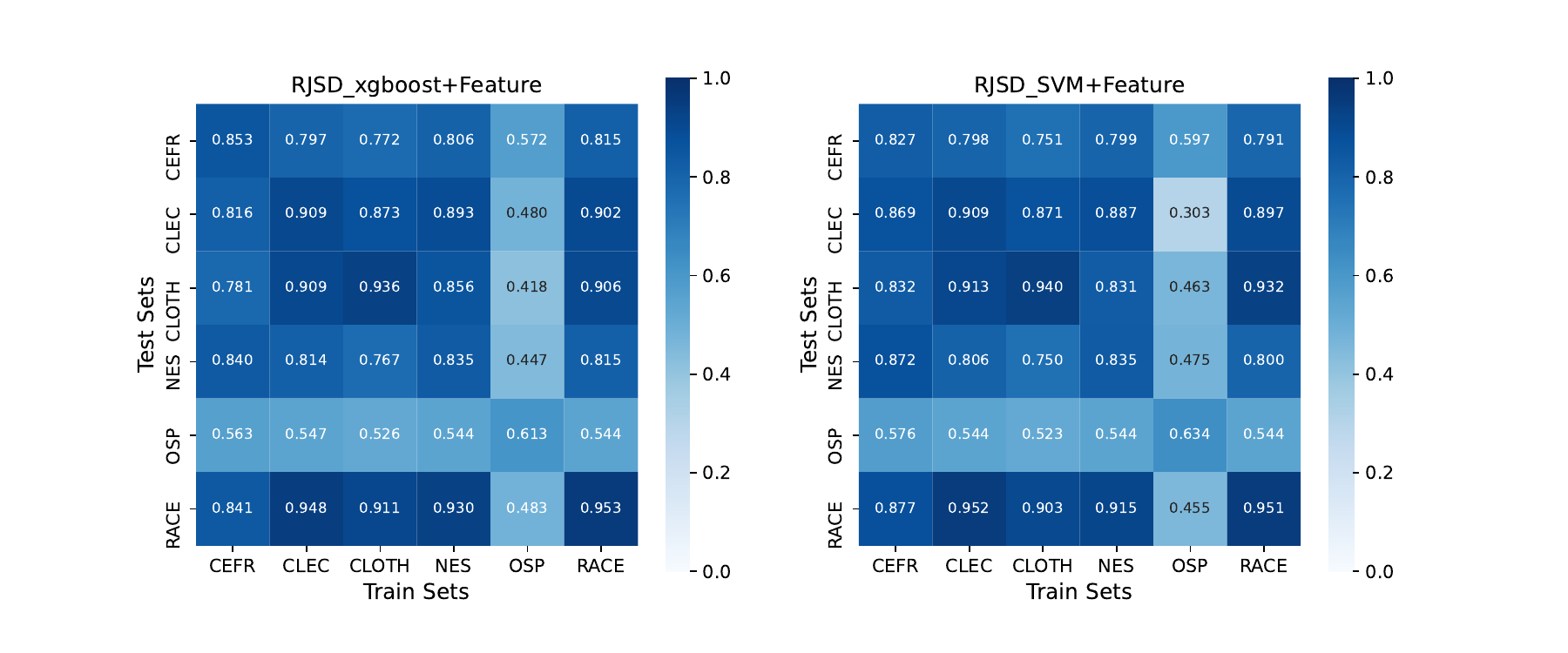}
    \caption{Compatibility Assessment Results of RJSD (ML+Feature)}
    \label{fig:RJSD (ML+Feature)}
\end{figure}

Figure \ref{fig:RJSD (ML+Feature)} shows the results of evaluating machine learning using linguistic features for prediction. The horizontal axis represents the source corpora used by the training model, and the vertical axis represents the target corpora used for prediction evaluation. The values in Figure \ref{fig:RJSD (ML+Feature)} are the compatibility assessment results using the RJSD method, with a value range from 0 to 1. The larger the result value, the higher the compatibility.

According to the comparison experimental results in Figure \ref{fig:RJSD (ML+Feature)}, it can be seen that the results obtained by the XGBoost and SVM methods are similar. When the RACE and CLEC corpora are used as the target corpora, the compatibility scores are higher compared to other corpora. For example, when using the XGBoost and SVM methods to train the CLEC corpus, the compatibility values for RACE reach the highest values of 0.948 and 0.952, respectively. On the other hand, when the OSP corpus is used as the target corpus for prediction evaluation, its compatibility score is relatively lower compared to other corpora. For example, by using SVM to train the source corpus CLOTH, the lowest value of the model result is 0.523.

Obviously, there is low compatibility shown on the RJSD measurement standard when the model trained on the OSP source corpus predicts the target corpus. The average results for XGBoost and SVM are 0.502 and 0.488, respectively. In contrast, the models trained on the CLEC and RACE source corpora have higher compatibility assessment results. The average results for XGBoost are 0.821 and 0.823, respectively, and the average results for SVM are 0.82 and 0.819, respectively.

\subsubsection{Compatibility Analysis of Using GloVe Word Vectors}
\begin{figure}[htb]
    \centering
    \includegraphics[width=8.5cm]{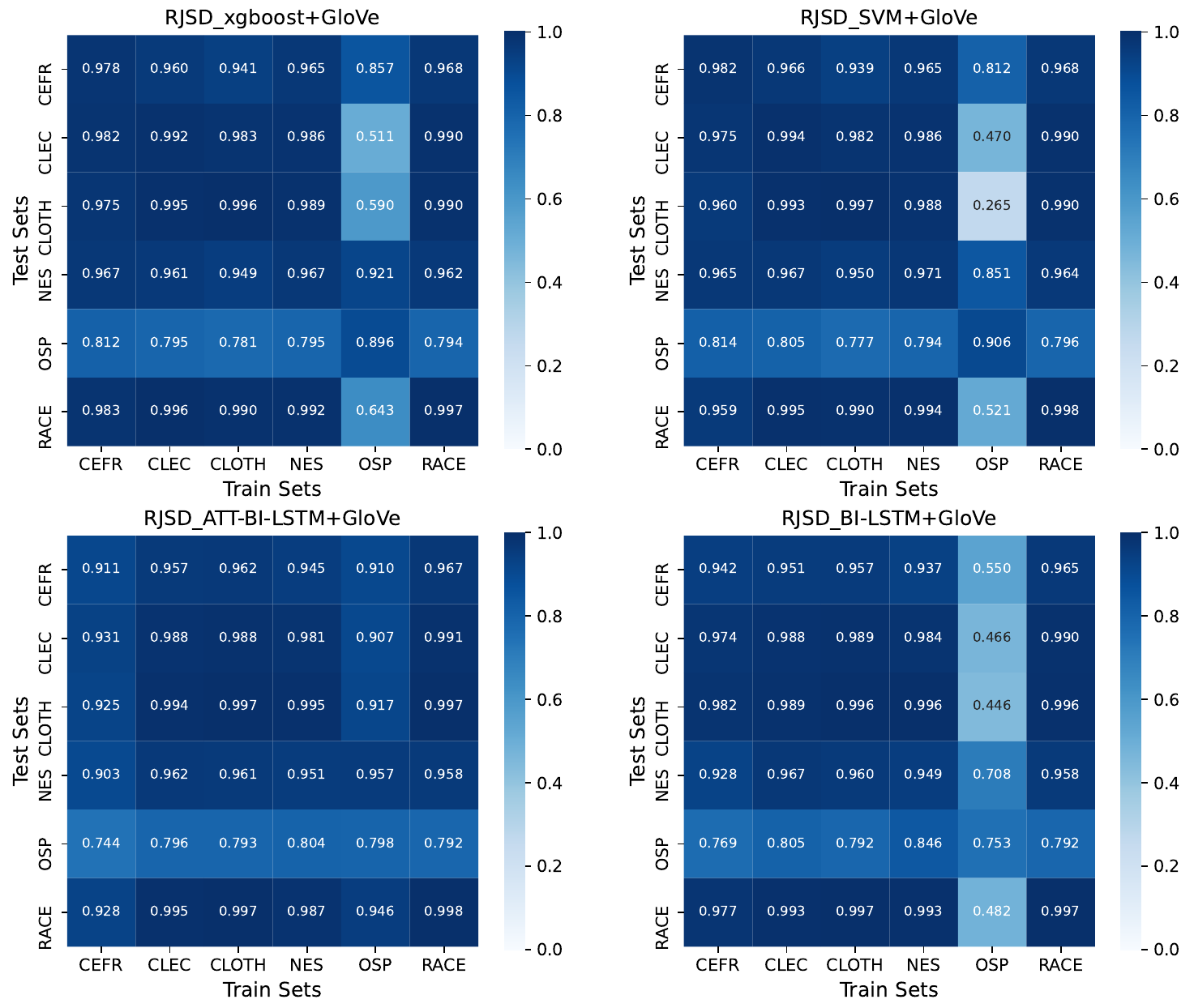}
    \caption{Compatibility Assessment Results of RJSD (ML/DL+GloVe)}
    \label{fig:RJSD (ML/DL+GloVe)}
\end{figure}

Figure \ref{fig:RJSD (ML/DL+GloVe)} shows the results of using GloVe word vectors for readability compatibility assessment by machine learning and deep learning methods. The four heatmap diagrams correspond to the results of compatibility assessment on the dataset using XGBoost, SVM, Attention-BiLSTM, and BiLSTM methods, respectively.

It can be seen from Figure \ref{fig:RJSD (ML/DL+GloVe)} that the compatibility assessment results of different target corpora are different. In particular, when the target corpus uses XGBoost and SVM methods to evaluate on different source corpora, the target corpus CEFR and NES have high compatibility with all corpora. In contrast, the compatibility between the target corpus OSP and other corpora is low. For example, the average values of XGBoost and SVM methods are 0.812 and 0.815, respectively. From the analysis of the source corpus, there are differences in compatibility between the OSP source corpus and the target corpus according to the RJSD measurement standard. For example, the compatibility between NES and the OSP source corpus reaches the highest value of 0.921 on the XGBoost method, while the compatibility between CLEC and the OSP source corpus is only 0.511 on the XGBoost method. This shows that the selection of the source corpus in the learning model is crucial to the performance of the algorithm.

When evaluated using Attention-BiLSTM and BiLSTM methods, they show the same level of compatibility as the XGBoost and SVM methods. Different from XGBoost and SVM methods, the compatibility of the target corpus with OSP as the source corpus is improved. The improvement in the Attention-BiLSTM method is obvious, and the average compatibility value reaches 0.906. This shows that the deep learning methods can improve the accuracy and compatibility of OSP corpus when evaluating the readability of cross-corpus text.

\subsubsection{Compatibility Analysis of Using Fusion Features}
\begin{figure}[htb]
    \centering
    \includegraphics[width=8.5cm]{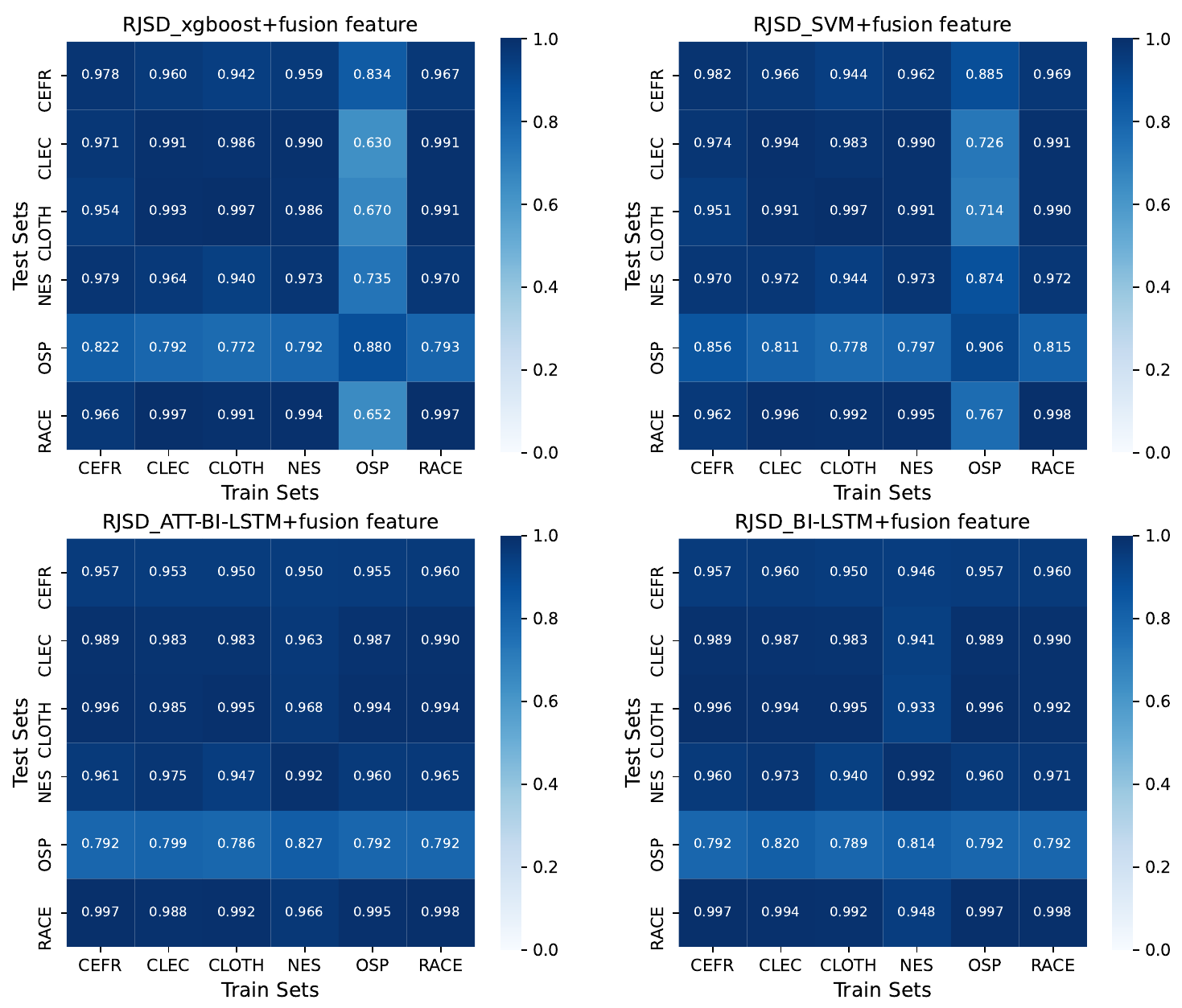}
    \caption{Compatibility Assessment Results of RJSD (ML/DL+fusion feature)}
    \label{fig:RJSD (ML/DL+fusion feature)}
\end{figure}

This experiment combines linguistic features with GloVe word vector representation as a new feature to train the model. Figure \ref{fig:RJSD (ML/DL+fusion feature)} shows the compatibility assessment results using fusion features according to the RJSD metric.

From the results shown in the figure, it can be observed that the performance using fusion features is similar to the previous two experiments. For example, the compatibility assessment results vary across different corpora. On XGBoost and SVM methods, there is lower compatibility between the target corpora CLEC, CLOTH, RACE and the OSP corpus. However, CEFR and NES corpora have better compatibility results with all corpora. When OSP is used as the source corpus, the cross-corpus compatibility results are not good. Similarly, when evaluated using Attention-BiLSTM or BiLSTM methods, it can be seen that the compatibility between the OSP corpus and all source corpora is relatively low. The average compatibility values were 0.798 and 0.800, respectively.

Different from the above two sets of experimental results, we found that when using the method of fusion features to train the model, the compatibility performance between different corpora was significantly improved. For example, when using the SVM+GloVe method to train a model on the OSP source corpus, the compatibility value between the CLOTH corpus and the OSP corpus was 0.265. However, when using the SVM+fusion feature method to train the model, the compatibility result is improved to 0.714. This shows that the fusion feature method has excellent performance in cross-corpus English text readability evaluation, and can effectively transfer learning for different data sets. The performance of Attention-BiLSTM and BiLSTM methods is more obvious. The combination of deep learning methods and fusion features makes the OSP source corpus obtain the same high compatibility as other corpora, with an average compatibility value of 0.947 and 0.949, respectively. This shows that the combination of deep learning method and fusion feature technology can improve the accuracy and compatibility of cross-corpus text readability assessment.

\begin{figure}[htp]
    \centering
    \includegraphics[width=8.5cm]{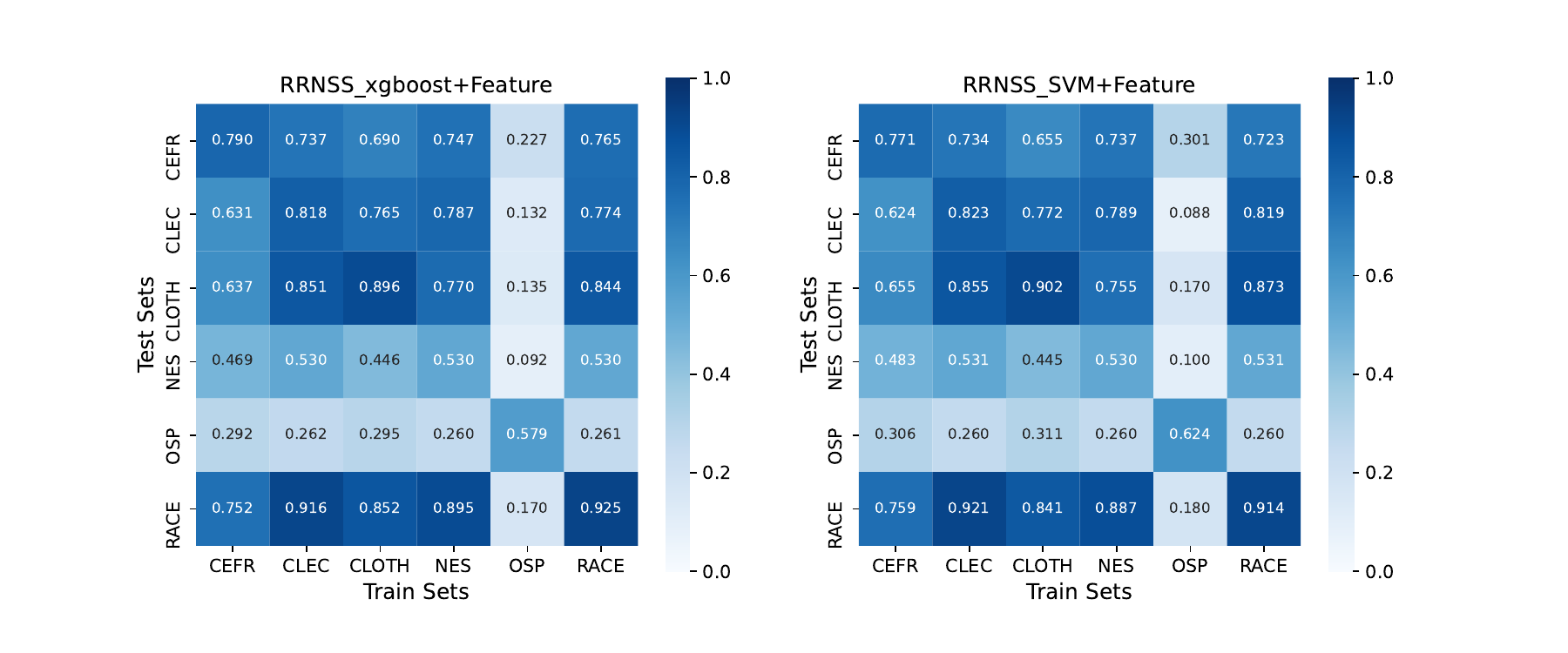}
    \caption{Compatibility Assessment Results of RRNSS (ML+Feature)}
    \label{fig:RRNSS (ML+Feature)}
\end{figure}

\begin{figure}[htp]
    \centering
    \includegraphics[width=8.5cm]{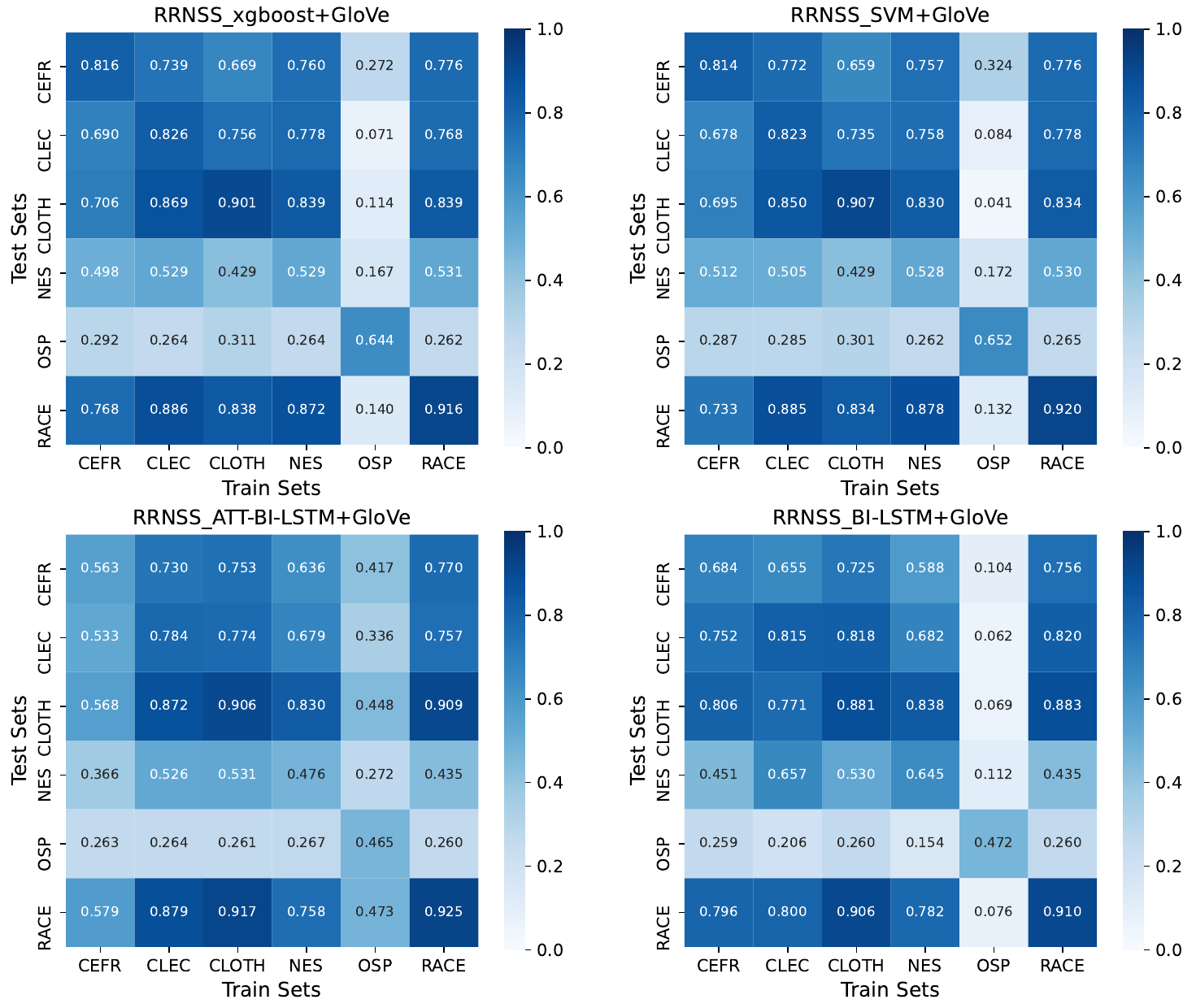}
    \caption{Compatibility Assessment Results of RRNSS (ML/DL+GloVe)}
    \label{fig:RRNSS (ML/DL+GloVe)}
\end{figure}

\begin{figure}[htp]
    \centering
    \includegraphics[width=8.5cm]{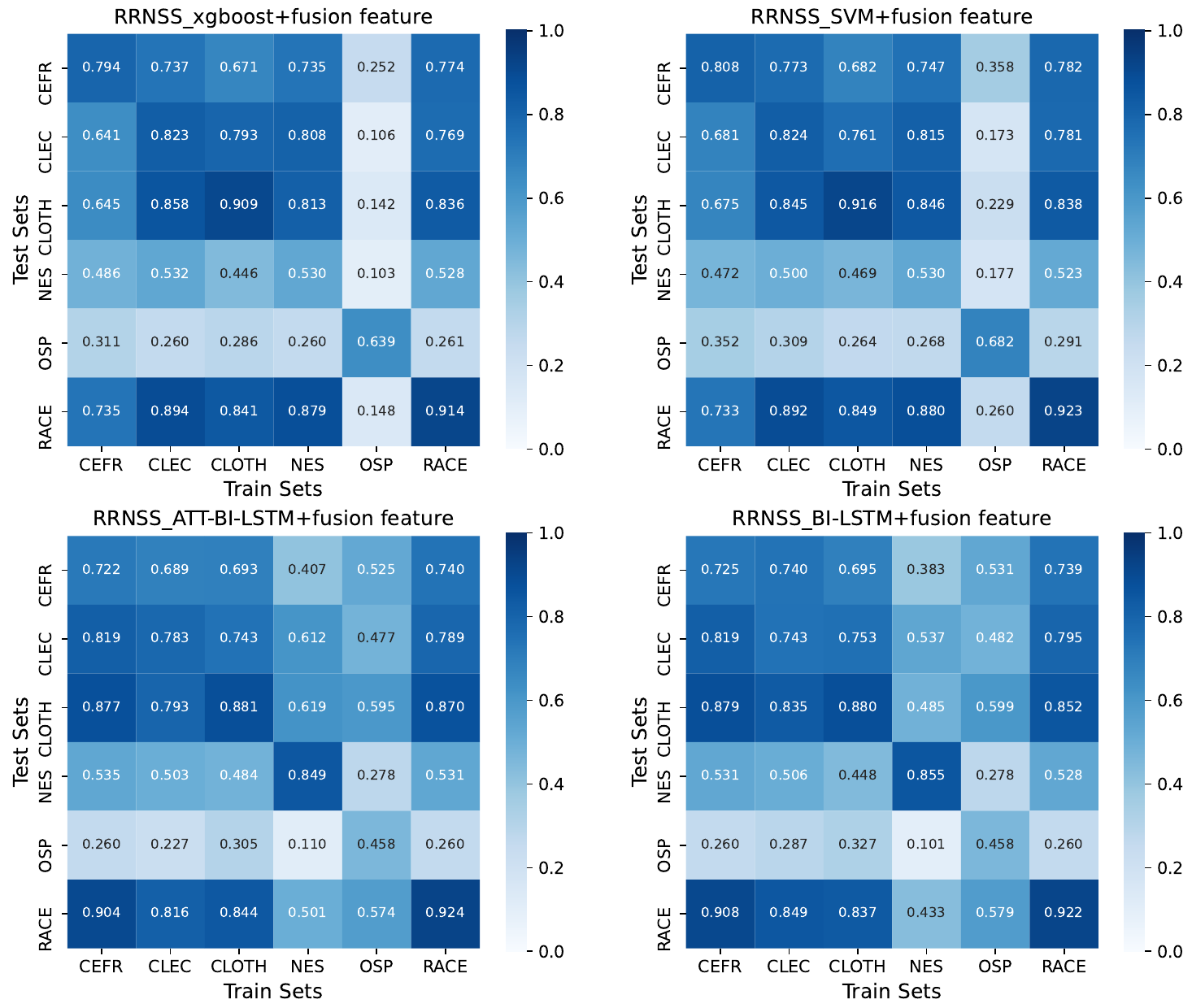}
    \caption{Compatibility Assessment Results of RRNSS (ML/DL+Fusion Feature)}
    \label{fig:RRNSS (ML/DL+Fusion Feature)}
\end{figure}

\begin{figure}[htp]
    \centering
    \includegraphics[width=8.5cm]{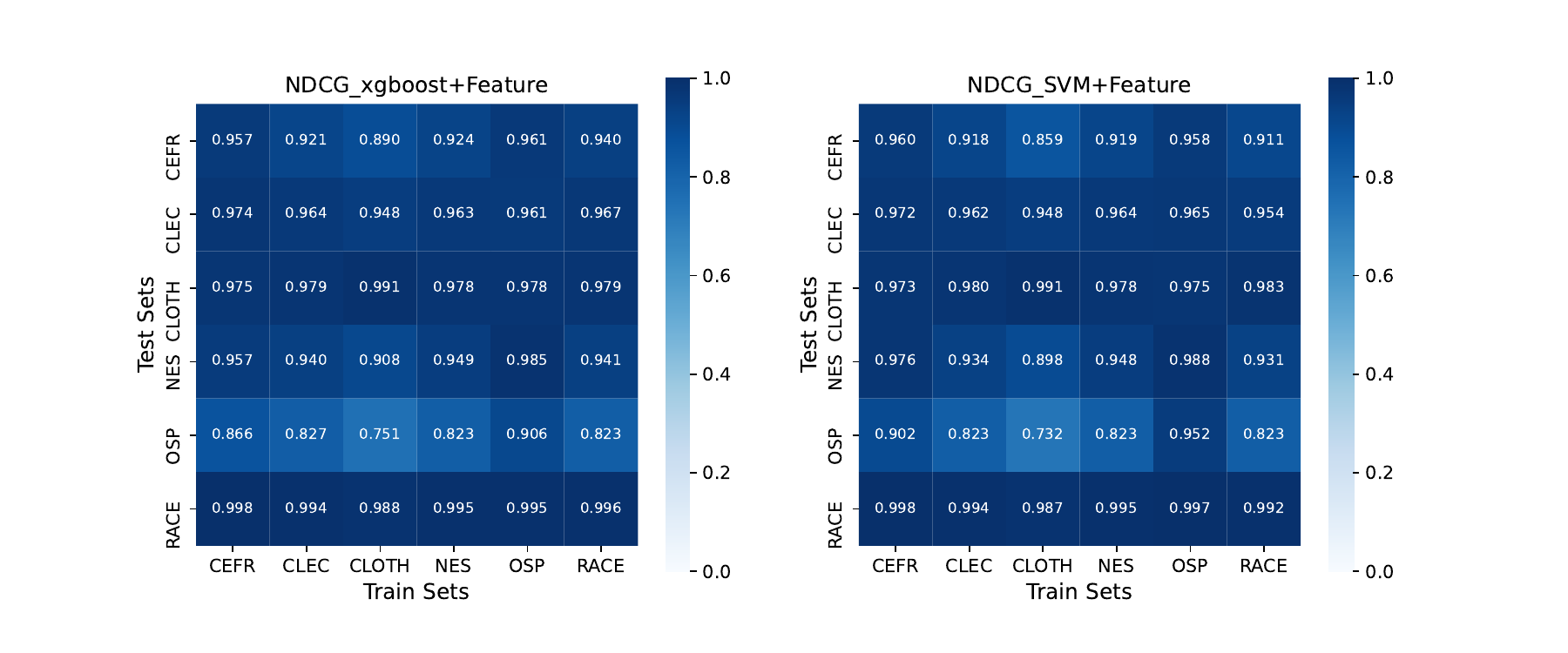}
    \caption{Compatibility Assessment Results of NDCG (ML+Feature)}
    \label{fig:NDCG (ML+Feature)}
\end{figure}

\begin{figure}[htp]
    \centering
    \includegraphics[width=8.5cm]{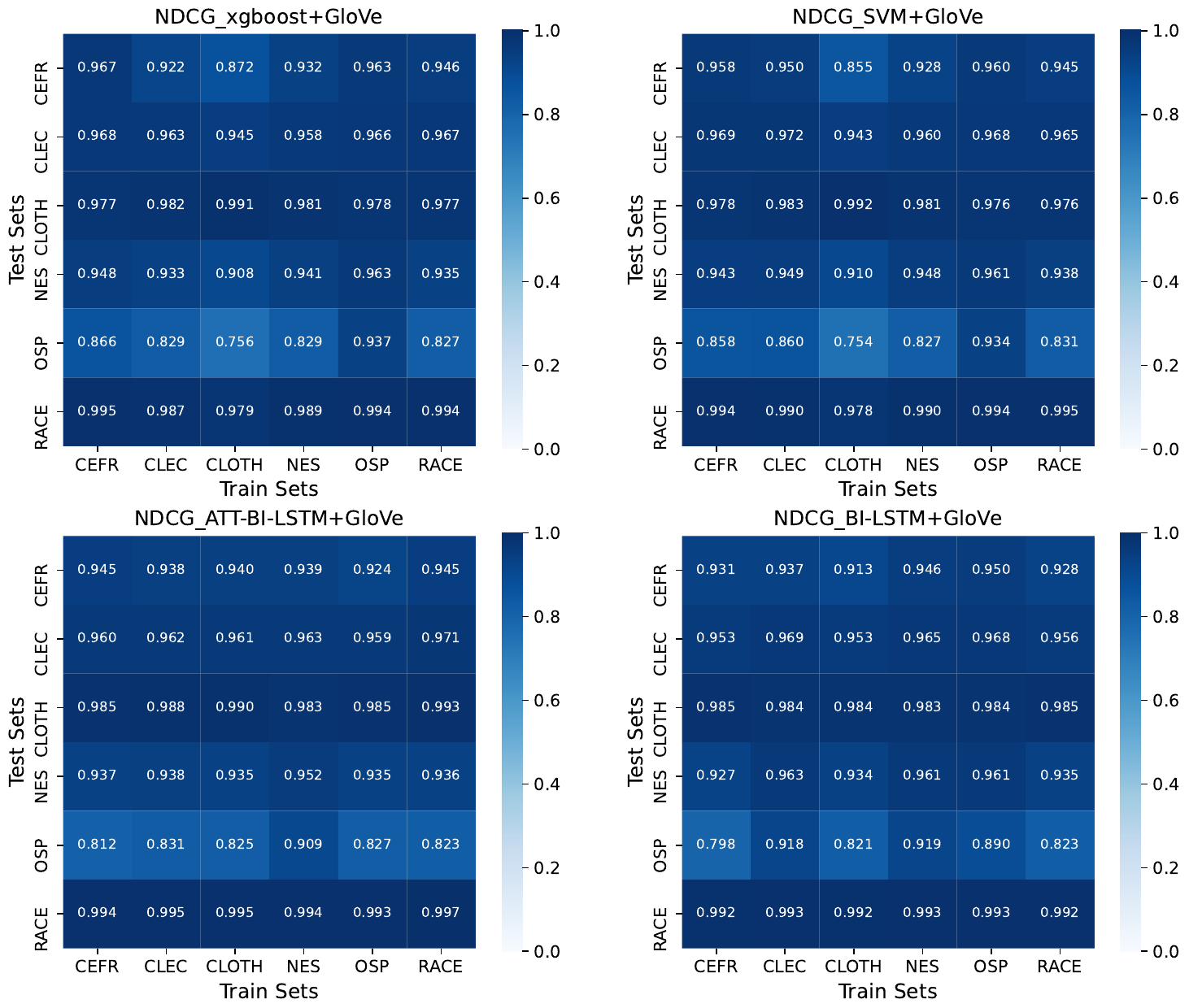}
    \caption{Compatibility Assessment Results of NDCG (ML/DL+GloVe)}
    \label{fig:NDCG (ML/DL+GloVe)}
\end{figure}

\begin{figure}[htp]
    \centering
    \includegraphics[width=8.5cm]{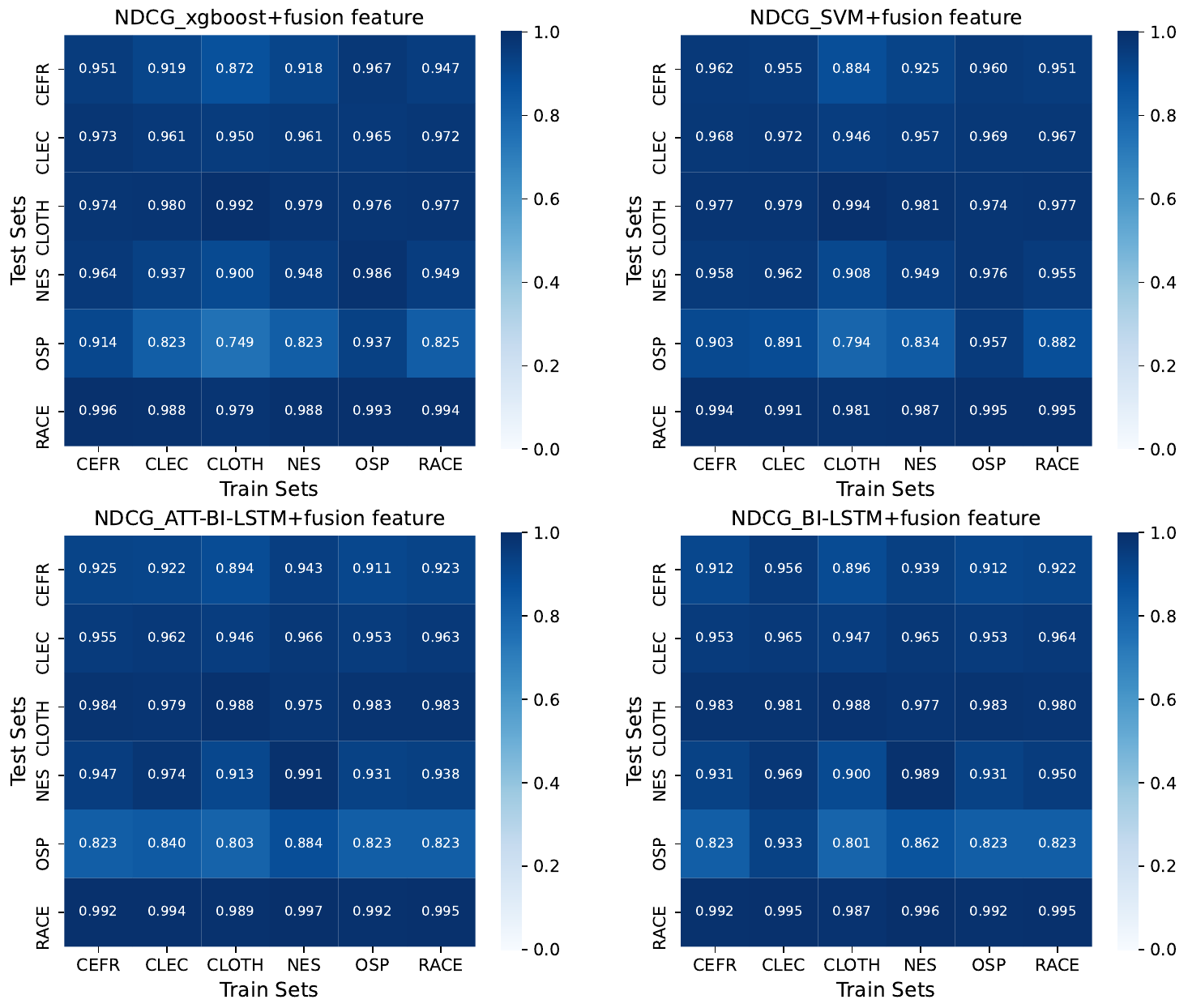}
    \caption{Compatibility Assessment Results of NDCG (ML/DL+Fusion Feature)}
    \label{fig:NDCG (ML/DL+Fusion Feature)}
\end{figure}

Similarly, in this paper, compatibility assessment experiments were conducted using the RRNSS and NDCG metrics. The results of RRNSS are shown in Figure \ref{fig:RRNSS (ML+Feature)}, Figure \ref{fig:RRNSS (ML/DL+GloVe)}, and Figure \ref{fig:RRNSS (ML/DL+Fusion Feature)}, respectively. The results of NDCG are shown in Figure \ref{fig:NDCG (ML+Feature)}, Figure \ref{fig:NDCG (ML/DL+GloVe)}, and Figure \ref{fig:NDCG (ML/DL+Fusion Feature)}, respectively. The measurement methods of different assessment methods are different, but similar conclusions can be drawn in the experiments. This verifies the robustness and reliability of our analysis results.

\subsection{The correlation between Compatibility Assessment metrics}
To further analyze the correlation between the results of the compatibility metrics, we conducted a correlation analysis of the results using three methods for measuring the compatibility of readability systems. In this analysis, we used the Pearson correlation coefficient to measure the correlation \cite{pearson1895vii}.

\begin{equation}
    r=\frac{n(\sum x y)-(\sum x)(\sum y)}{\sqrt{[n\sum x^{2}-\left(\sum x\right)^{2}][n\sum y^{2}-\left(\sum y\right)^{2}]}}
\end{equation}

where, $r=$ Pearson Coefficient, $n=$ number of pairs of the stock, $\sum x y=$ sum of products of the paired stocks, $\sum x=$ sum of the x scores, $\sum y=$ sum of the y scores, $\sum x^{2}=$ sum of the squared x scores, and $\sum y^{2}=$ sum of the squared y scores.

\begin{table}[htp]
\caption{Results of the correlation between metrics of all datasets}
\label{tab:correlation between metrics}
\centering
\footnotesize
\begin{tabular}{llll}
\toprule[1.5pt]
 & RJSD & RRNSS & NDCG \\
\midrule
RJSD     & 1.000 (0.0) & 0.761 (3.99E-69) & 0.317 (7.43E-10)\\
RRNSS    & 0.761 (3.99E-69) & 1.000 (0.0) & 0.488 (6.47E-23)\\
NDCG     & 0.317 (7.43E-10) & 0.488 (6.47E-23) & 1.000 (0.0)\\
\bottomrule[1.5pt]
\end{tabular}
\end{table}

In Table \ref{tab:correlation between metrics}, we present the Pearson correlation results and corresponding p-values for the three Compatibility Assessment metrics. It can be seen from the table that the self-correlation analysis for each metric resulted in 1. The correlation results along the diagonal are the same, with a confidence level of less than 0.01. Meanwhile, the correlation between the RJSD metric and the RRNSS metric is 0.761, the correlation between the RJSD metric and the NDCG metric is 0.317, and the correlation between the RRNSS metric and the NDCG metric is 0.488. Thus, the results suggest that the RJSD metric is more highly correlated with the RRNSS metric than with the NDCG metric.

\section{Conclusion}

In this work, we present CRCA to evaluate the compatibility of cross-corpus difficulty systems. Through our experiments on popular benchmark corpora, we observe that the RACE, CLOTH, CLEC, CEFR, and NES corpora exhibit high compatibility with each other, while the OSP corpus demonstrates low compatibility with all of the other corpora. Furthermore, the analysis of the source corpus reveals variations in compatibility when paired with different target corpora, highlighting the critical role of source corpus selection in algorithm performance. This finding is particularly evident in the SVM+GloVe method, where the compatibility between OSP and NES is measured at RJSD = 0.851, whereas it is RJSD = 0.265 when evaluating compatibility with CLOTH.

Our comparison of cross-corpus compatibility assessment results encompasses two aspects: feature combination and classification methods. We find that the applicability of classification methods for assessing text difficulty compatibility depends on the chosen feature combination. The combination of Attention-BiLSTM, BiLSTM methods, and fusion features significantly enhances the compatibility of the OSP source corpus. Compared to GloVe word vector representation, the average compatibility values increase by 0.214 and 0.137, respectively. This demonstrates the superior performance of deep learning methods and fusion features in cross-corpus English text readability assessment.

Moreover, we employ RJSD, RRNSS, and NDCG as evaluation metrics for compatibility results in our experiments. Despite their distinct perspectives, these metrics consistently yield similar overall outcomes, confirming the robustness and reliability of CRCA. Our work not only facilitates comparisons in corpus selection but also provides experimental evidence for validating findings in the context of natural language processing research and practical applications.

\bibliographystyle{IEEEtran}
\bibliography{IEEEabrv,cas-refs}

\appendices
\section{}
\label{appendix:index score}

\renewcommand{\arraystretch}{1.2} 
\begin{table}[htp]
\caption{Readability formula scores of datasets}
\centering
\footnotesize
\resizebox{8.5cm}{!}{
\begin{tabular}{llllllllll}
\toprule[1.5pt]
Class        & ARI    & FKGL   & GFI    & SMOG  & CLI   & LIX    & RIX   \\
\midrule
\multicolumn{8}{c}{CEFR}                                                 \\
\hline
A1           & 2.57   & 2.83   & 7.01   & 7.25  & 4.24  & 21.79  & 1.19  \\
A2           & 3.81   & 3.55   & 7.03   & 7.14  & 5.31  & 24.98  & 1.62  \\
B1           & 8.10   & 7.06   & 10.68  & 9.87  & 8.51  & 36.40  & 3.31  \\
B2           & 7.89   & 7.04   & 10.75  & 9.77  & 8.26  & 36.09  & 3.26  \\
C1           & 10.56  & 9.24   & 13.14  & 11.66 & 10.63 & 43.55  & 4.67  \\
C2           & 11.26  & 10.00  & 14.21  & 12.27 & 10.16 & 44.23  & 4.94  \\
\hline
\multicolumn{8}{c}{CLEC}                                                 \\
\hline
ST2          & 51.25  & 40.63  & 44.63  & 20.62 & 12.07 & 122.40 & 22.08 \\
ST3          & 65.69  & 52.04  & 56.60  & 26.80 & 14.34 & 151.91 & 32.51 \\
ST4          & 60.05  & 47.46  & 52.25  & 24.19 & 14.13 & 140.05 & 27.44 \\
ST5          & 90.88  & 72.00  & 76.35  & 29.57 & 15.40 & 203.17 & 49.48 \\
ST6          & 151.10 & 119.72 & 125.63 & 43.99 & 16.06 & 324.29 & 88.71 \\
\hline
\multicolumn{8}{c}{CLOTH}                                                \\
\hline
middle       & 3.87   & 3.08   & 7.58   & 7.30  & 4.46  & 25.86  & 1.77  \\
high         & 7.26   & 5.74   & 10.66  & 9.05  & 5.71  & 33.48  & 2.79  \\
\hline
\multicolumn{8}{c}{NES}                                                  \\
\hline
2nd grade    & 2.85   & 3.09   & 5.64   & 6.75  & 5.82  & 23.23  & 1.20  \\
3rd grade    & 4.45   & 4.43   & 6.90   & 7.62  & 7.49  & 27.50  & 1.69  \\
4th grade    & 5.44   & 5.27   & 7.87   & 8.26  & 8.08  & 30.08  & 2.11  \\
5th grade    & 6.93   & 6.60   & 9.18   & 9.16  & 9.21  & 34.10  & 2.73  \\
6th grade    & 8.20   & 7.74   & 10.38  & 9.92  & 9.69  & 37.19  & 3.34  \\
7th grade    & 9.46   & 8.89   & 11.47  & 10.68 & 10.57 & 40.53  & 3.97  \\
8th grade    & 10.55  & 9.70   & 12.39  & 11.18 & 10.63 & 42.72  & 4.50  \\
9th grade    & 11.65  & 10.75  & 13.38  & 11.89 & 11.39 & 45.61  & 5.12  \\
10th grade   & 11.66  & 10.69  & 13.50  & 11.91 & 11.11 & 45.51  & 5.13  \\
12th grade   & 13.09  & 11.87  & 14.69  & 12.60 & 11.37 & 48.54  & 5.93  \\
\hline
\multicolumn{8}{c}{OSP}                                                  \\
\hline
beginner     & 8.56   & 7.55   & 11.06  & 10.29 & 9.24  & 36.96  & 3.42  \\
intermediate & 10.60  & 9.41   & 13.03  & 11.65 & 10.29 & 42.17  & 4.47  \\
advanced     & 12.14  & 10.67  & 14.39  & 12.51 & 10.87 & 45.99  & 5.33  \\
\hline
\multicolumn{8}{c}{RACE}                                                 \\
\hline
middle       & 5.09   & 5.17   & 8.13   & 7.85  & 5.92  & 28.26  & 2.17  \\
high         & 10.37  & 9.57   & 12.54  & 10.88 & 9.27  & 41.33  & 4.50 \\
\bottomrule[1.5pt]
\end{tabular}
}

\end{table}
\newpage
\section{}
\label{appendix:Distribution}

\begin{figure}[htp]
    \centering
    \includegraphics[width=8.5cm]{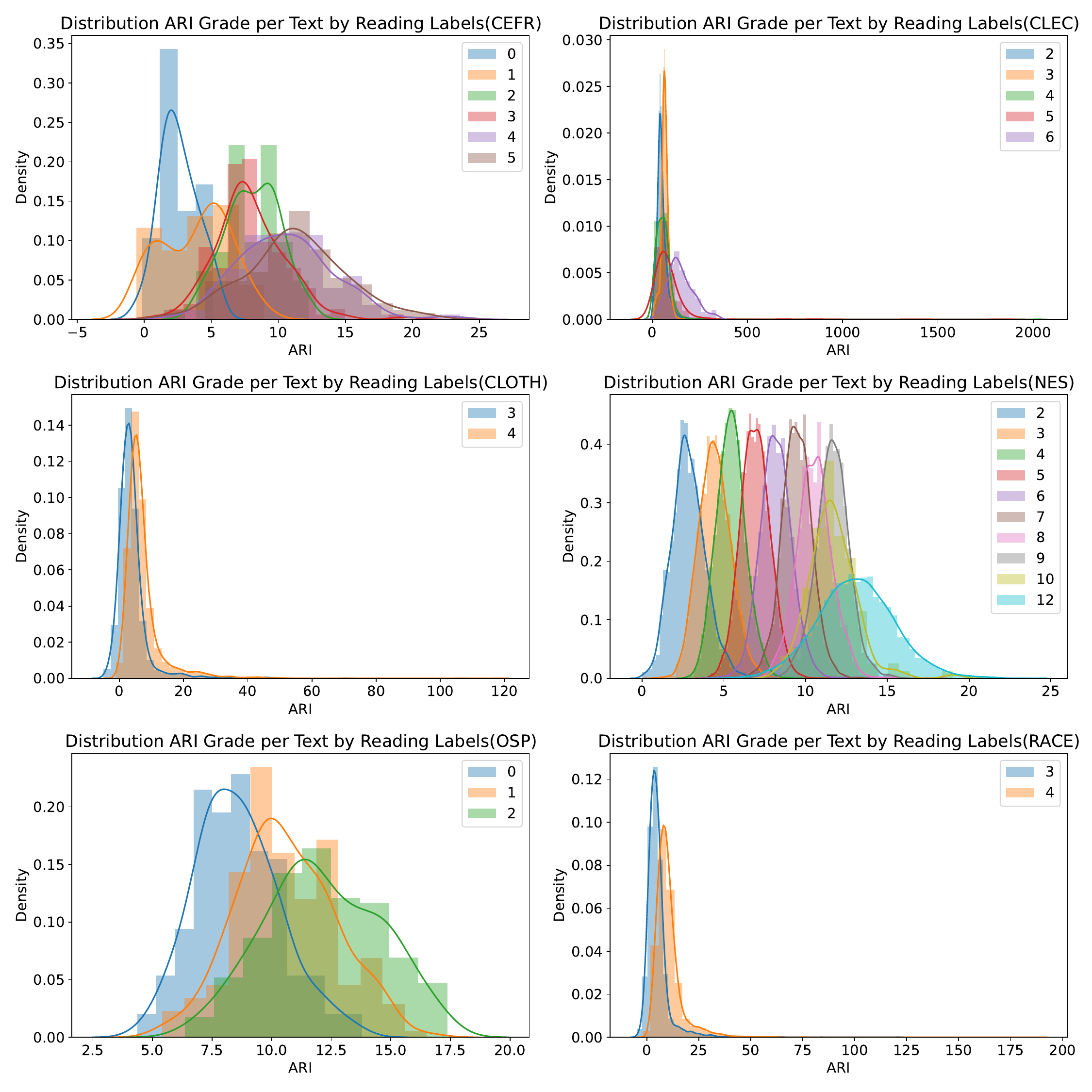}
    \caption{Distribution ARI Grade per Text by Reading Labels}
\end{figure}

\begin{figure}[htp]
    \centering    
    \includegraphics[width=8.5cm]{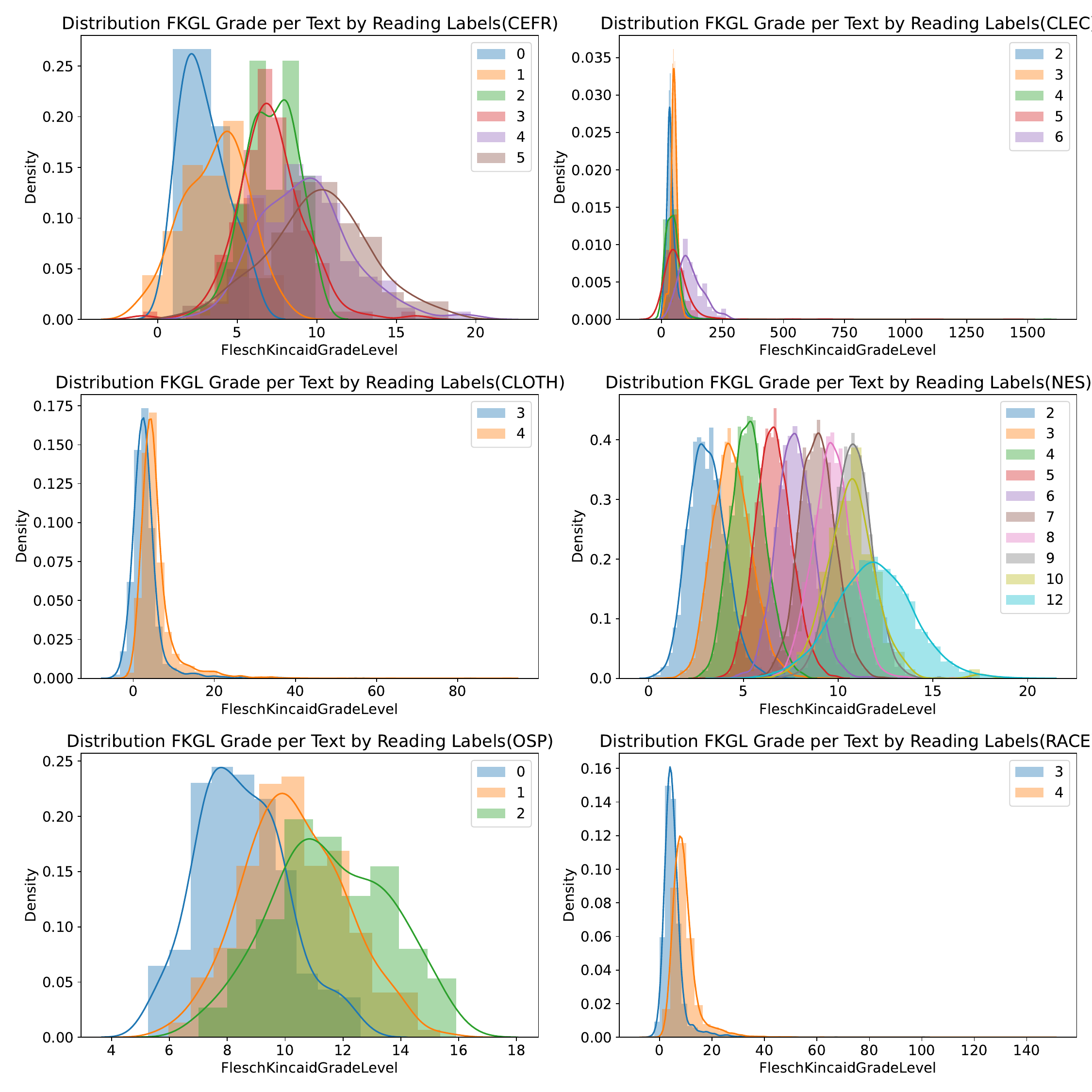}
    \caption{Distribution FKGL Grade per Text by Reading Labels}
\end{figure}

\begin{figure}[htp]
    \centering
    \includegraphics[width=8.5cm]{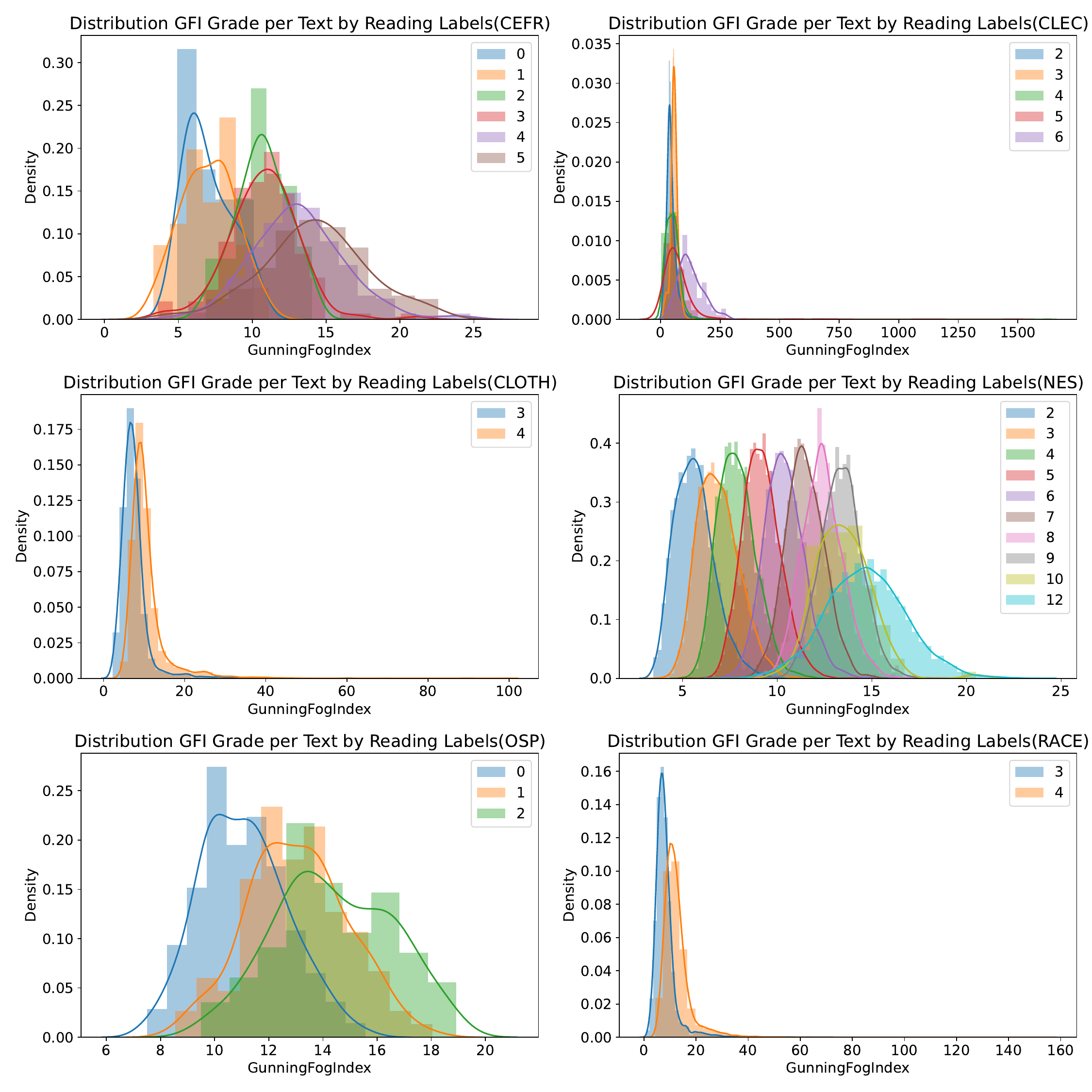}
    \caption{Distribution GFI Grade per Text by Reading Labels}
\end{figure}

\begin{figure}[htp]
    \centering
    \includegraphics[width=8.5cm]{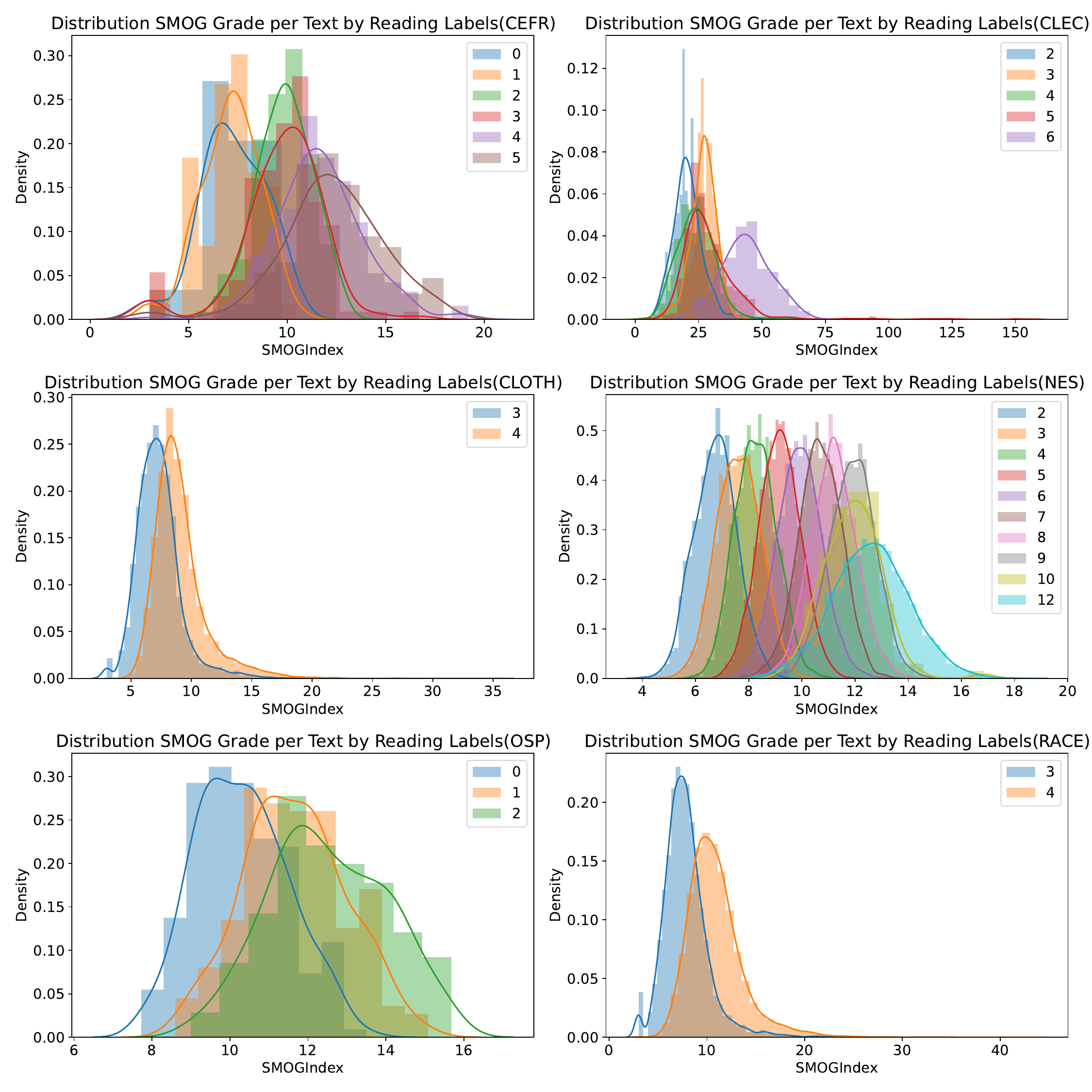}
    \caption{Distribution SMOG Grade per Text by Reading Labels}
\end{figure}

\begin{figure}[htp]
    \centering
    \includegraphics[width=8.5cm]{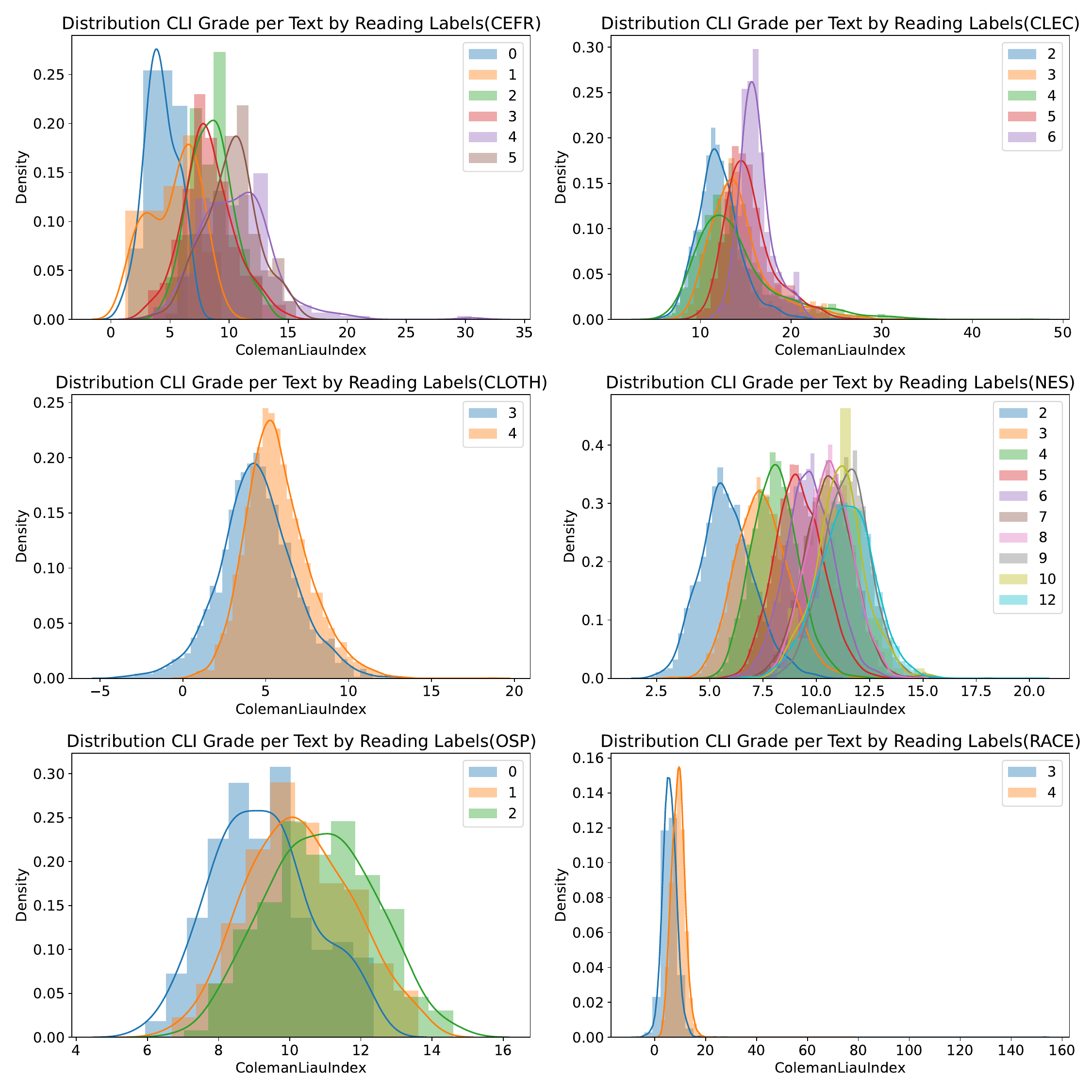}
    \caption{Distribution CLI Grade per Text by Reading Labels}
\end{figure}

\begin{figure}[htp]
    \centering    
    \includegraphics[width=8.5cm]{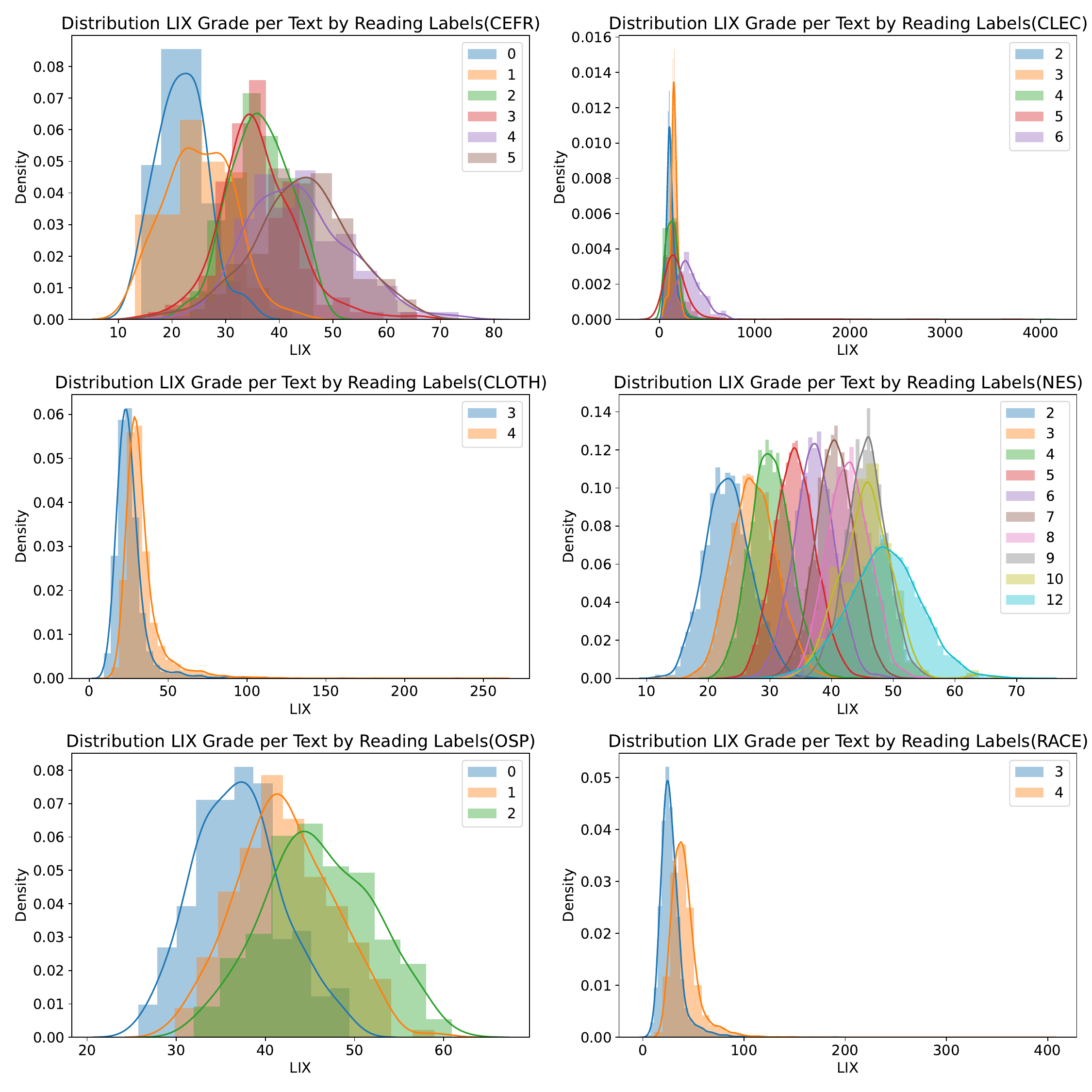}
    \caption{Distribution LIX Grade per Text by Reading Labels}
\end{figure}

\begin{figure}[htp]
    \centering
    \includegraphics[width=8.5cm]{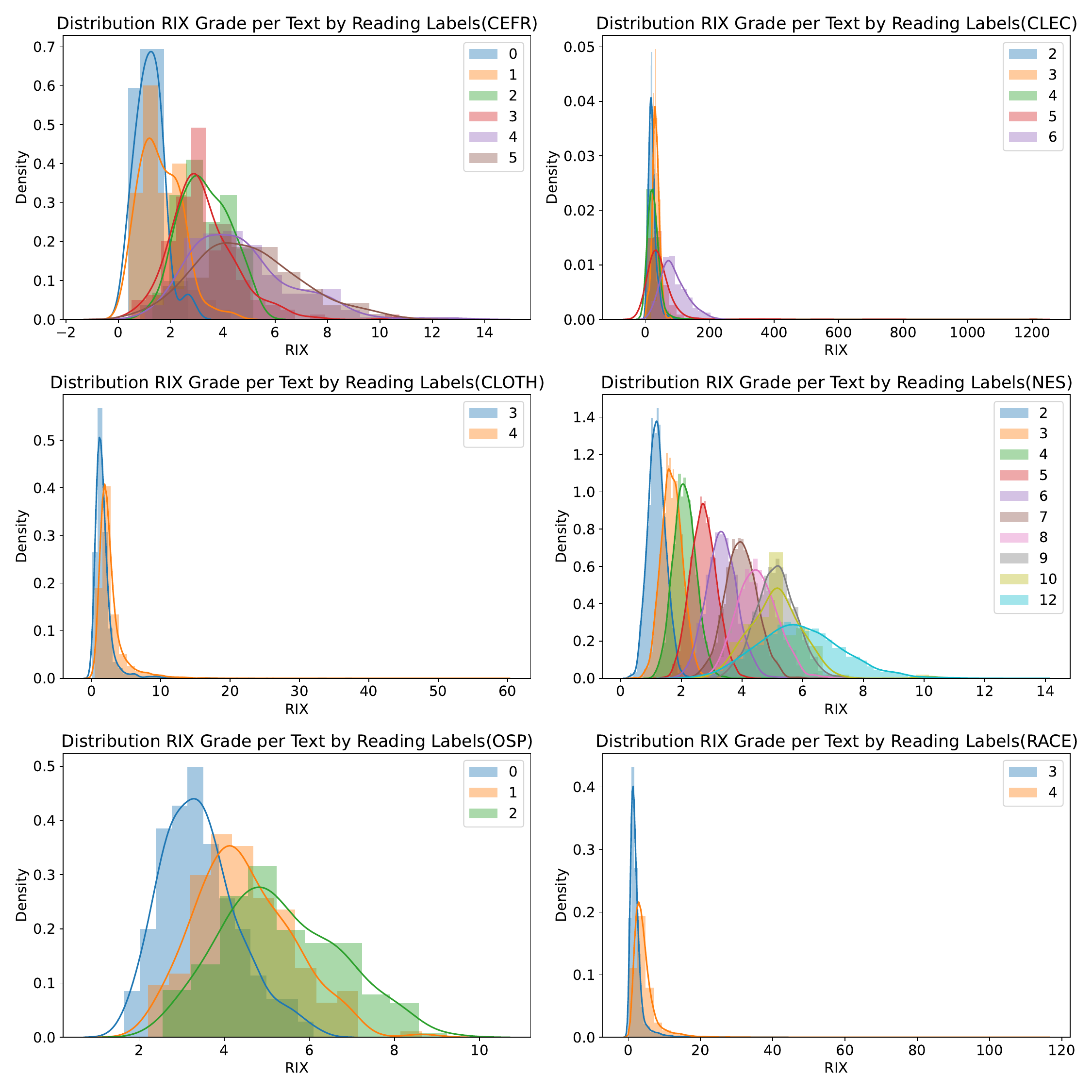}
    \caption{Distribution RIX Grade per Text by Reading Labels}
\end{figure}


\newpage

\section{}
\label{appendix:Feature}

\begin{table}[h]
\centering
\caption{Feature Information}
\renewcommand{\arraystretch}{1.2}
\label{tab:Information}
\resizebox{9cm}{!}{
\begin{tabular}{ccl}
\toprule[1.5pt]
Count & Code & Description \\
\hline
\multicolumn{3}{c}{Lexical Features}  \\
\hline
0     & ALPW & average letters per word                                                   \\
1     & ASPW & average syllables per word                                                  \\
2     & CWP  & percentage of complex words                                                \\
3     &  DWP  & percentage of difficult words                                                 \\
4     & LWP  & percentage of long words                                                     \\
\hline
\multicolumn{3}{c}{Syntactic Features}                                                      \\
\hline
5     & ANPS & average length of noun phrases                                               \\
6     & APPS & average length of prepositional phrases                                      \\
7     & APT  & average length of parse tree                                                 \\
8     & AVPS & average length of verb phrases                                               \\
9     & AWPS & average number of words per sentence                                         \\
10    & LSP  & percentage of long sentences                                                 \\
11    & NPS  & average number of noun phrases per sentence                                  \\
12    & PPS  & average number of prepositional phrases per sentence                         \\
13    & SPS  & average number of subordinate clauses per sentence                           \\
14    & SQS  & average number of special subordinate clause per sentence \\
15    & VPS  & average number of verb phrases per sentence                                  \\
\hline
\multicolumn{3}{c}{Grammatical Features}                                                    \\
\hline
16    & CoP  & percentage of conjunctions                                                   \\
17    & CP   & percentage of commas\\
18    & NP   & percentage of nouns                                                          \\
19    & PNP  & percentage of proper nouns                                                   \\
20    & PP   & percentage of pronouns                                                       \\         \bottomrule[1.5pt]                                             
\end{tabular}
}
\end{table}

\section{}
\label{appendix:Model Performance}

\begin{table}[htp]
\renewcommand{\arraystretch}{1}
\caption{Model Performance}
\resizebox{8.5cm}{!}{
\begin{tabular}{lccccc}
\toprule
\quad   & Method     & Accuracy & Precision & Recall & F1 \\
\midrule



\multicolumn{6}{c}{CEFR} \\
\hline
\multirow{2}{*}{Feature} 
& XGBoost & 0.416    & 0.207     & 0.290  & 0.222 \\
& SVM        & 0.415    & 0.291     & 0.363  & 0.278 \\
\hline
\multirow{4}{*}{GloVe}         
& XGBoost    & 0.431    & 0.283     & 0.325  & 0.287 \\
& SVM        & 0.547    & 0.552     & 0.512  & 0.525 \\
& BiLSTM     & 0.309    & 0.232     & 0.205  & 0.154 \\
& Att-BiLSTM & 0.300    & 0.085     & 0.141  & 0.102 \\
\hline
\multirow{4}{*}{Fusion Feature} 
& XGBoost    & 0.387    & 0.492     & 0.371  & 0.385 \\
& SVM        & 0.518    & 0.475     & 0.545  & 0.503 \\
& BiLSTM     & 0.373    & 0.896     & 0.167  & 0.091 \\
& Att-BiLSTM & 0.327    & 0.598     & 0.155  & 0.104 \\
\hline
\multicolumn{6}{c}{CLEC}\\
\toprule
\multirow{2}{*}{Feature}      
& XGBoost    & 0.416    & 0.207     & 0.290  & 0.222 \\
& SVM        & 0.415    & 0.291     & 0.363  & 0.278 \\
\hline
\multirow{4}{*}{GloVe}        
& XGBoost    & 0.431    & 0.309     & 0.376  & 0.292 \\
& SVM        & 0.528    & 0.229     & 0.428  & 0.326 \\
& BiLSTM     & 0.394    & 0.497     & 0.363  & 0.328 \\
& Att-BiLSTM & 0.411    & 0.243     & 0.362  & 0.260 \\
\hline
\multirow{4}{*}{Fusion Feature} 
& XGBoost    & 0.401    & 0.287     & 0.351  & 0.267 \\
& SVM        & 0.528    & 0.298     & 0.455  & 0.351 \\
& BiLSTM     & 0.478    & 0.464     & 0.421  & 0.378 \\
& Att-BiLSTM & 0.406    & 0.703     & 0.362  & 0.260 \\
\hline
\multicolumn{6}{c}{CLOTH}\\
\toprule
\multirow{2}{*}{Feature}       
& XGBoost    & 0.732    & 0.726     & 0.727  & 0.727 \\
& SVM        & 0.750    & 0.501     & 0.489  & 0.492 \\
\hline
\multirow{4}{*}{GloVe}         
& XGBoost    & 0.746    & 0.740     & 0.739  & 0.739 \\
& SVM        & 0.773    & 0.768     & 0.769  & 0.769 \\
& BiLSTM     & 0.593    & 0.711     & 0.514  & 0.400 \\
& Att-BiLSTM & 0.751    & 0.829     & 0.705  & 0.705 \\
\hline
\multirow{4}{*}{Fusion Feature} 
& XGBoost    & 0.783    & 0.779     & 0.776  & 0.777 \\
& SVM        & 0.818    & 0.813     & 0.815  & 0.814 \\
& BiLSTM     & 0.834    & 0.836     & 0.822  & 0.827 \\
& Att-BiLSTM & 0.831    & 0.831     & 0.819  & 0.823 \\
\hline
\multicolumn{6}{c}{NES}\\
\toprule
\multirow{2}{*}{Feature}       
& XGBoost    & 0.272    & 0.170     & 0.239  & 0.179 \\
& SVM        & 0.275    & 0.176     & 0.241  & 0.181 \\
\hline
\multirow{4}{*}{GloVe}         
& XGBoost    & 0.157    & 0.100     & 0.127  & 0.084 \\
& SVM        & 0.452    & 0.366     & 0.353  & 0.343 \\
& BiLSTM     & 0.271    & 0.091     & 0.195  & 0.105 \\
& Att-BiLSTM & 0.195    & 0.157     & 0.124  & 0.068 \\
\hline
\multirow{4}{*}{Fusion Feature} 
& XGBoost    & 0.274    & 0.191     & 0.263  & 0.198 \\
& SVM        & 0.286    & 0.198     & 0.277  & 0.208 \\
& BiLSTM     & 0.513    & 0.690     & 0.359  & 0.281 \\
& Att-BiLSTM & 0.516    & 0.665     & 0.373  & 0.329 \\
\hline
\multicolumn{6}{c}{OSP}\\
\toprule
\multirow{2}{*}{Feature}       
& XGBoost    & 0.605    & 0.641     & 0.605  & 0.616 \\
& SVM        & 0.561    & 0.474     & 0.421  & 0.431 \\
\hline
\multirow{4}{*}{GloVe}         
& XGBoost    & 0.605    & 0.719     & 0.605  & 0.613 \\
& SVM        & 0.623    & 0.612     & 0.623  & 0.617 \\
& BiLSTM     & 0.484    & 0.467     & 0.474  & 0.446 \\
& Att-BiLSTM & 0.264    & 0.500     & 0.269  & 0.223 \\
\hline
\multirow{4}{*}{Fusion Feature} 
& XGBoost    & 0.614    & 0.663     & 0.614  & 0.623 \\
& SVM        & 0.728    & 0.733     & 0.728  & 0.729 \\
& BiLSTM     & 0.363    & 0.669     & 0.364  & 0.238 \\
& Att-BiLSTM & 0.429    & 0.754     & 0.411  & 0.308 \\
\hline
\multicolumn{6}{c}{RACE}\\
\toprule
\multirow{2}{*}{Feature}       
& XGBoost    & 0.837    & 0.527     & 0.511  & 0.518 \\
& SVM        & 0.775    & 0.856     & 0.563  & 0.548 \\
\hline
\multirow{4}{*}{GloVe}         
& XGBoost    & 0.819    & 0.819     & 0.815  & 0.817 \\
& SVM        & 0.837    & 0.836     & 0.835  & 0.836 \\
& BiLSTM     & 0.751    & 0.715     & 0.537  & 0.506 \\
& Att-BiLSTM & 0.850    & 0.816     & 0.779  & 0.795 \\
\hline
\multirow{4}{*}{Fusion Feature} 
& XGBoost    & 0.850    & 0.823     & 0.759  & 0.782 \\
& SVM        & 0.846    & 0.845     & 0.844  & 0.845 \\
& BiLSTM     & 0.822    & 0.778     & 0.737  & 0.752 \\
& Att-BiLSTM & 0.850    & 0.816     & 0.778  & 0.794\\
\hline
\end{tabular}
}
\end{table}


\pagebreak

\begin{IEEEbiography}[{\includegraphics[width=1in,height=1.25in,clip,keepaspectratio]{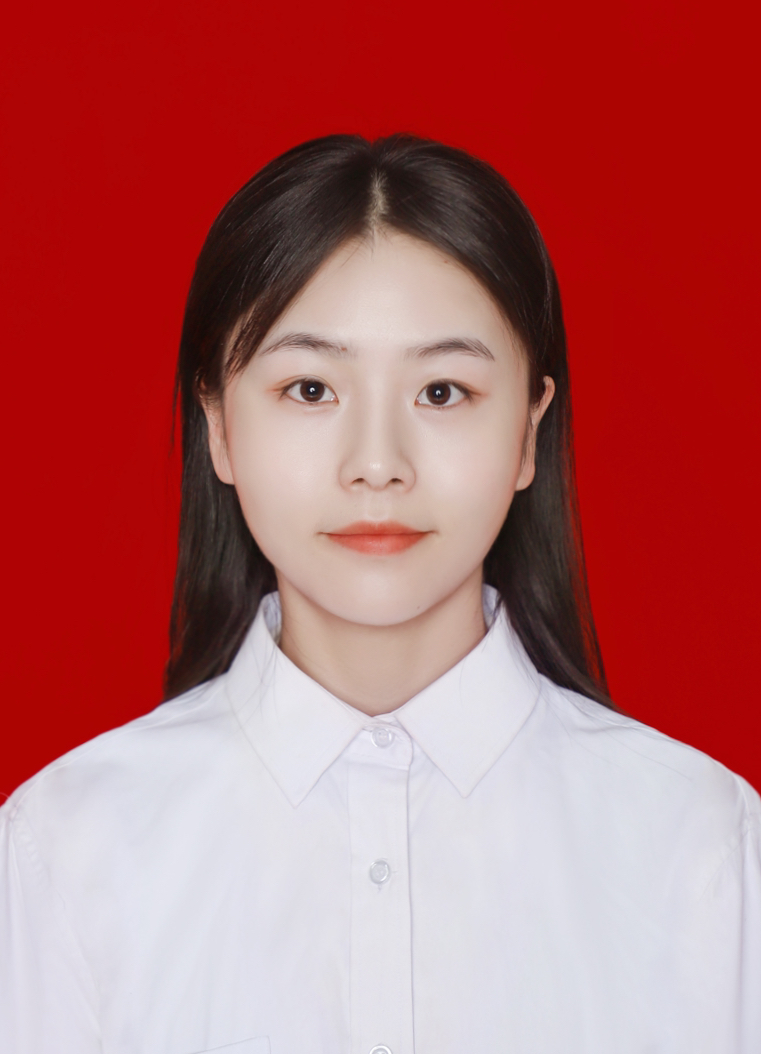}}]{Zhenzhen Li} was born in Henan Province. She received her B.E. degree from Xinyang Agriculture and Forestry University,  Xinyang, China, in 2021. She is currently pursuing her M.E. degree at Guangzhou University, Guangzhou, China. Her research interests primarily focus on data mining, machine learning, natural language processing, and deep learning.
\end{IEEEbiography}

\begin{IEEEbiography}[{\includegraphics[width=1in,height=1.25in,clip,keepaspectratio]{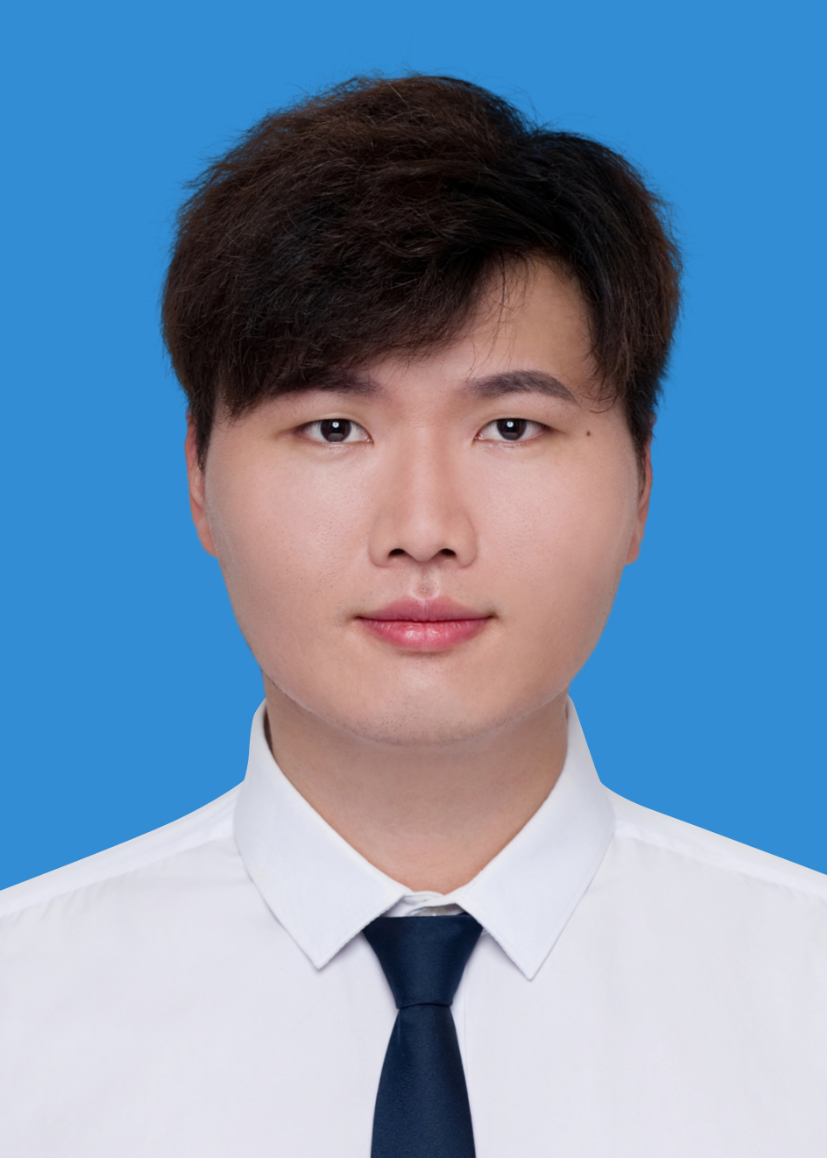}}]{Han Ding} was born in Jiangsu Province. He received his B.E. degree from Nanjing Xiaozhuang University, Nanjing, China, in 2020. He is currently pursuing his M.E. degree at Guangzhou University, Guangzhou,China. He research interests primarily focus on data mining, text readability, natural language processing, and learning to rank.

\end{IEEEbiography}

\begin{IEEEbiography}[{\includegraphics[width=1in,height=1.25in,clip,keepaspectratio]{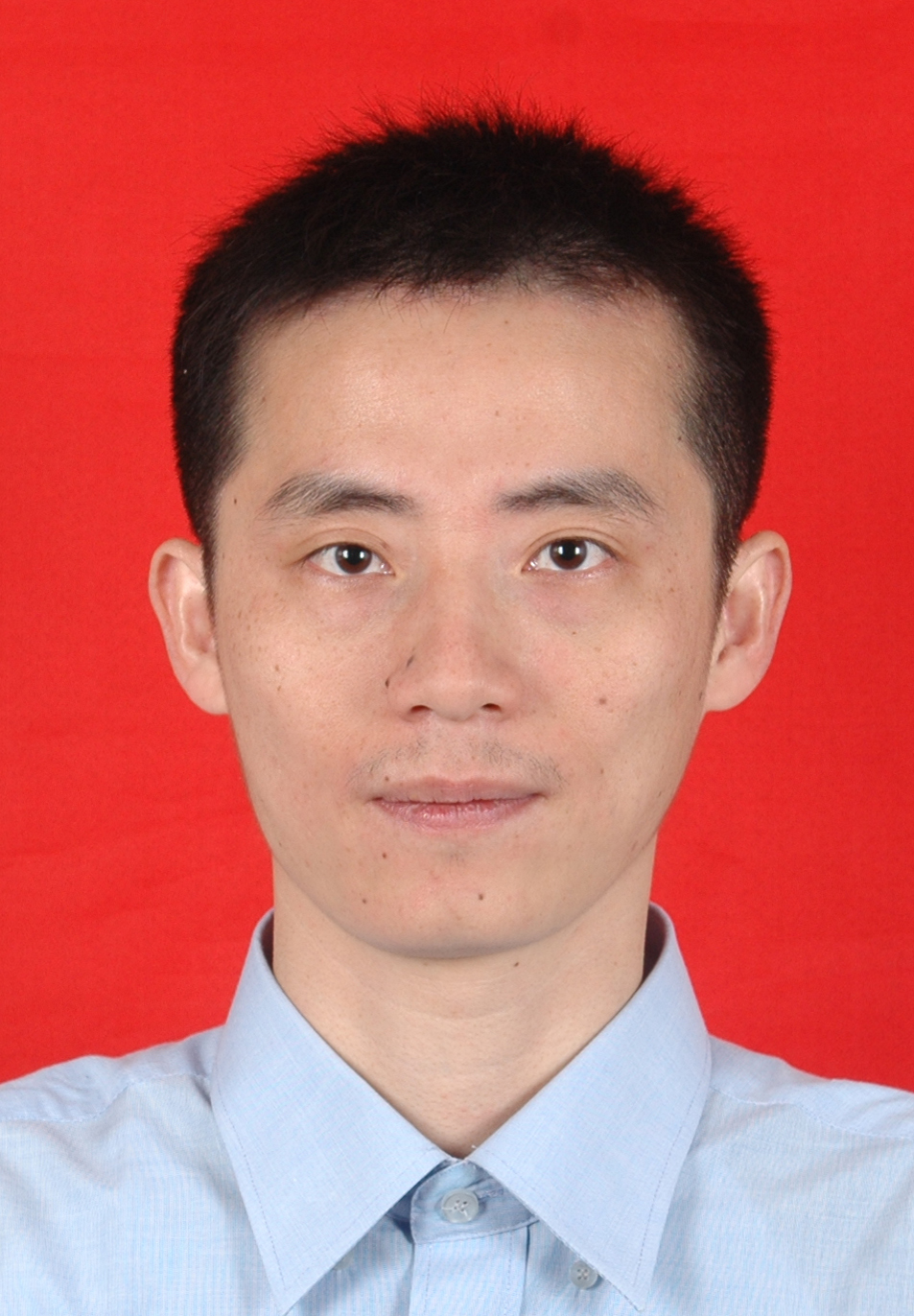}}]{Shaohong Zhang} is an associate professor in the Department of Computer Science at Guangzhou University. He was a postdoctoral fellow in the Department of Computer Science, City University of Hong Kong. He received the PhD degree from Department of Computer Science, City University of Hong Kong. His research interests include pattern recognition, data mining, and bioinformatics. He has published more than 60 articles.
\end{IEEEbiography}

\EOD

\end{document}